\documentclass[preprint,final,12pt]{elsarticle}

\usepackage{hyperref}

\journal{Pattern Recognition}

\usepackage{hyphenat}
\usepackage[cmex10]{amsmath}
\usepackage{amsfonts}
\usepackage{capt-of}
\usepackage{xcolor}
\usepackage{xspace}
\usepackage{subfig}

\newcommand{\DSMARQUES}{\textsc{S-Marques}\xspace}
\newcommand{\DISRI}{\textsc{Isri-OCR}\xspace}
\newcommand{\DSISRI}{\textsc{S-Isri-OCR}\xspace}
\newcommand{\DCDIP}{\textsc{Cdip}\xspace}
\newcommand{\DSCDIP}{\textsc{S-Cdip}\xspace}
\newcommand{\fig}[1]{Figure \ref{#1}}
\newcommand{\sect}[1]{Section \ref{#1}}
\newcommand{\eq}[1]{Equation \ref{#1}}

\graphicspath{{figs/}}

\begin{document}

\begin{frontmatter}
\title{Self-supervised Deep Reconstruction of Mixed Strip-shredded Text Documents}




\author[ifes,ufes]{Thiago M. Paix\~ao\corref{mycorrespondingauthor}}
\cortext[mycorrespondingauthor]{Corresponding author}
\ead{paixao@gmail.com}

\author[ufes]{Rodrigo F. Berriel}
\ead{rfberriel@inf.ufes.br}

\author[ufes]{Maria C. S. Boeres}
\ead{boeres@inf.ufes.br}

\author[ets]{Alessandro L. Koerich}
\ead{alessandro.koerich@etsmtl.ca}

\author[ufes]{Claudine Badue}
\ead{claudine@lcad.inf.ufes.br}

\author[ufes]{Alberto F. De Souza}
\ead{alberto@lcad.inf.ufes.br}

\author[ufes]{Thiago Oliveira-Santos}
\ead{todsantos@lcad.inf.ufes.br}

\address[ifes]{Instituto Federal do Esp\'irito Santo (IFES), ES-010 Km-6.5, Manguinhos, Serra - ES, Brazil, 29173-087}
\address[ufes]{Universidade Federal do Esp\'irito Santo (UFES), Av. Fernando Ferrari, 514 - Goiabeiras, Vit\'oria - ES, Brazil, 29075-910}
\address[ets]{\'Ecole de Technologie Sup\'erieure (\'ETS), Universit\'e du Qu\'ebec, 1100 Notre-Dame West Street, Montreal, QC H3C 1K3, Canada}

\begin{abstract}
{The reconstruction of shredded documents consists of coherently arranging fragments of paper (shreds) to recover the original document(s). A great challenge in computational reconstruction is to properly evaluate the compatibility between the shreds. While traditional pixel-based approaches are not robust to real shredding, more sophisticated solutions compromise significantly time performance. The solution presented in this work extends our previous deep learning method for single-page reconstruction to a more realistic/complex scenario: the reconstruction of several mixed shredded documents at once. In our approach, the compatibility evaluation is modeled as a two-class (valid or invalid) pattern recognition problem. The model is trained in a self-supervised manner on samples extracted from simulated-shredded documents, which obviates manual annotation. Experimental results on three datasets -- including a new collection of 100 strip-shredded documents produced for this work -- have shown that the proposed method outperforms the competing ones on complex scenarios, achieving accuracy superior to $90\%$.}
\end{abstract}

\begin{keyword}
Deep Learning, Self-supervised Learning, Fully Convolutional Neural Networks, Document Reconstruction, Forensics, Optimization Search.
\end{keyword}

\end{frontmatter}

\section{Introduction}
\label{sec:introduction}


{Pattern recognition (PR) plays a significant role in image-based forensics. There is a wide range of interesting topics, such as person identification based on biometrics \cite{cheng2019deep}, postmortem identification \cite{gomez20183d}, and the recent challenge of DeepFake detection \cite{li2020cvpr}. In particular, PR-based similarity analysis can enable forensic document examination (FDE) to solve forgery detection \cite{qureshi2019hyperspectral}, author identification \cite{he2019deep}, signature authenticity verification \cite{soleimani2016deep}, and other kinds of problems.}

{Complementarity (or compatibility) between patterns is a related concept to similarity (present in some of the aforementioned applications) that can also benefit from PR \cite{townshend2019end,ostertag2020matching}. For instance, shape analysis \cite{paixaotifs2019} and machine learning \cite{perl2011} have been used to analyze patterns (e.g., lines, shapes, characters) that emerge when two fragments of papers (shreds) are placed side-by-side. Such an ability is helpful to reconstruct manually and mechanically-shredded documents, where the purpose is to arrange the shreds -- likewise in a jigsaw puzzle -- to recover the original document(s). In the FDE context, the reconstruction may be imperative for further analysis if the information contained in such documents is essential to decide criminal and civil cases.}



{The automatic and digital reconstruction of shredded documents is desirable because the traditional (manual) approach is potentially damaging to the paper due to the continuous direct contact with the shreds, besides being a slow and tedious process for humans. Computational solutions} typically involve an optimization process guided by an objective function that quantifies the global fitting (compatibility) of the shreds. This function can be computed directly from the entire image representing a reconstruction  \cite{badawy2018discrete} or derived from the compatibility between shreds (pairwise compatibility), being the latter more common.

In this direction, a recent work \cite{paixaotifs2019} verified the lack of robust pairwise compatibility measures addressing {real mechanically-shredded documents (i.e., cut by paper shredder machines)} with low color information and predominant textual content. To fill this gap, the authors {explored PR by using} similarity between characters' shape as criteria for compatibility determination rather than traditional edge pixel-wise correlation, as used in \cite{marques2013,gong2016,chen2018high}.
The characters' shape-based approach yielded better accuracy than the state-of-the-art in the reconstruction of strip-shredded documents (i.e., cut only in the longitudinal direction) at that time, however, with a significantly higher computational cost that limits the application for a large number of documents. Furthermore, the efficacy of the method strongly depends on the proper segmentation of text areas in the shreds, which may be challenging for more complex documents mixing text and pictorial elements (e.g., tables, diagrams, photos).

In this work, deep learning is leveraged to enable faster and more robust reconstruction of strip-shredded documents in more realistic scenarios. In a segmentation-free approach, the fitting of the shreds is modeled as a two-class (valid or invalid) PR problem. For this purpose, a convolutional neural network (CNN) is trained with local (small) samples extracted around the cutting sections of digitally-cut documents, i.e., digital documents submitted to simulated shredding. Besides facilitating data acquisition, the simulated process also provides the ground-truth order of the shreds for free. This enables the self-supervised learning of the model, in which class labels are determined directly from adjacency information of the shreds. A preliminary investigation of this approach \cite{paixao2018deep} has indicated the potential of using simulated data to reconstruct real-shredded documents in constrained scenarios, such as single-page reconstruction and intra-database evaluation. The first scenario assumes that the shreds are previously separated by pages so that the reconstruction system should deal with one page at a time (limited to $30$ shreds per page). Note that this is far from a real case scenario with multiple documents to be reconstructed, but gives an idea of the method potential. On its turn, intra-database evaluation consists in using the same document collection to provide both training and test sets, which reduces the variation between the two sets.


This work extends the preliminary study \cite{paixao2018deep} by addressing the more realistic and challenging task of reconstructing, at once, several shredded documents with heterogeneous content. The proposed evaluation explores cross-database scenarios, i.e., training and testing in different collections of documents. As an additional contribution, a new dataset with $100$ real strip-shredded documents (totaling $2{,}292$ shreds) of heterogeneous types was created for more extensive experimentation and released to the community to overcome the lack of publicly available collections representing real scenarios. This new dataset alone comprises more documents (100) than the full collection used in \cite{paixao2018deep} (80). Moreover, our dataset is significantly more heterogeneous since it comprises 10 different document categories (e.g., handwritten documents and forms), which is not common in the literature on reconstruction of shredded documents. Such diversity enables confirming the generalization and robustness of the proposed method.

Other specific contributions of this work include: (i) a more in-depth description of the technical details; (ii) an extended discussion of the state-of-the-art; (iii) an ablation study to analyze the sensitivity of the method to some key parameters. Experiments were conducted on the two collections of shredded documents used in \cite{paixao2018deep}, as well as on the newly-introduced dataset. Results have shown that our deep learning approach outperformed the competing methods, being capable of reconstructing $100$ mixed shredded documents ($2{,}292$ shreds) with accuracy superior to $90\%$, which brings the state-of-the-art of the document reconstruction problem to another level.

The remainder of the text is organized as follows. The next section presents the related work. \sect{sec:proposed} describes the proposed reconstruction system. The experimental methodology and the obtained results are, respectively, in Sections \ref{sec:experimental} and \ref{sec:results}. Finally, conclusion and future works are drawn in \sect{sec:conclusion}.
\section{Related Work}
\label{sec:related_work}


The literature on reconstruction of shredded documents has mostly focused on improving the optimization search process, relegating the image-based problem of verifying shreds' compatibility to the background. Most works are restricted to the reconstruction of simulated-shredded documents \cite{morandell2008,prandtstetter2008,sleit2013, phienthrakul2015,gong2016}, and, in this situation, the criticalness of the compatibility evaluation step in the reconstruction pipeline is masked, giving rise to potentially misleading conclusions. Therefore, this review focuses on different approaches to assess compatibility between shreds, which is the main contribution of this work.

Inspired by the jigsaw-puzzle solving problem, most literature on reconstruction explores low-level features for compatibility evaluation. The most naive approach in this context is to apply distance metrics (e.g., Hamming, Euclidean, Canberra, Manhattan) on the raw pixels of opposite boundaries of two shreds \cite{smet2005,skeoch2006,marques2013,chen2017a,chen2019solution}.
Some of these methods rely on the very edge \cite{skeoch2006,marques2013}, being more sensitive to the corruption of the shreds' extremities caused by the mechanical cut. To alleviate this, Marques and Freitas \cite{marques2013} suggest removing some border pixels, which, in practice, results in limited improvement. Additionally, different color spaces have been investigated (RGB \cite{smet2005}, HSV 
\cite{skeoch2006,marques2013}, gray-scale \cite{chen2017a}) without great success for text documents due to their poor chromatic information.

More sophisticated compatibility measures were designed to solve image puzzles with rectangular tiles, such as prediction-based \cite{pomeranz2011} measure. In the context of document shreds, Andal\'o et al. \cite{andalo2017} proposed a modified version of the measure proposed by Pomeranz et al. \cite{pomeranz2011}, reaching near 100\% of accuracy for simulated shredding with documents from ISRI-Tk OCR dataset \cite{nartker2005}, a collection of images commonly used to assess optical character recognition (OCR) software. Nevertheless,
\cite{paixaotifs2019,paixao2018deep} demonstrated that the accuracy of their method decreases dramatically when dealing with real-shredded documents of the same image collection.

Some authors designed compatibility measures focusing on text documents \cite{balme2007,morandell2008,ranca2013}.
Balme \cite{balme2007} and Morandell \cite{morandell2008} addressed the problem of vertical misalignment between pixels around the cutting section, i.e., the area near the touching edges of two adjacent shreds. Both of them adopt binary image representation given the black-and-white appearance of the text documents. Balme's measure is used in several works \cite{prandtstetter2008,gong2016,chen2018high}, and consists of a weighted pixel correlation intended to mitigate the misalignment issue, whereas Morandell \cite{morandell2008} quantifies the degree of misalignment between corresponding text lines (``black'' pixels) as a measure of compatibility. In an unsupervised approach, Ranca \cite{ranca2013} proposed learning the expected arrangement of pixels around the cutting section using information from the pixels inside the shreds. The best results were achieved with a simple probabilistic model, although they have also evaluated, unsuccessfully, feed-forward neural networks. Their experiments were also limited, given that they were carried out only on simulated-shredded data. Text-line detection was exploited in \cite{lin2012,sleit2013,pohler2015}, however, these methods struggle with ambiguities typically found in common text documents.

At a higher level of abstraction, compatibility can be assessed by exploring the matching degree of fragmented characters around the cutting section \cite{perl2011,phienthrakul2015,xing2017a,xing2017b,paixaotifs2019}. The continuity of character strokes was used as a matching criterion in \cite{phienthrakul2015,xing2017a}. In such an approach, the reconstruction accuracy strongly relies on the vertical alignment of the shreds and the image quality around the shreds' boundaries.
Alternatively, learning-based matching has also been proposed in the literature \cite{perl2011,xing2017b,paixaotifs2019}. Matching in \cite{perl2011} leverages OCR based on keypoint features. OCR tends to work well for general text recognition, but its application on corrupted characters is quite unstable, which turns it into a drawback of this formulation. Instead of identifying symbols, Xing et al. \cite{xing2017b} proposed a learning model to identify valid combinations of symbols (restricted to the Chinese language) based on structural features. The work of Paix\~ao et al. \cite{paixaotifs2019},
introduced in the previous section, analyzes the types of symbol combinations based on their shapes. Both \cite{xing2017b,paixaotifs2019} depends on segmenting text information from shreds.

In recent work, Liang and Li \cite{liang2019reassembling} proposed the \emph{word path metric}, which combines pixel- and character-level information (low-level metrics) with word-level information (high-level metric). A central procedure in their method is sampling candidate sequences and applying OCR for word recognition to improve pair compatibility estimation. Despite reporting accuracy comparable to \cite{paixao2018deep} (the preliminary investigation on deep learning-based reconstruction), their validation relies on solely three real-shredded instances. For two of them, those which are up to 39 shreds, their method achieved accuracy above $70\%$, while the third instance, with 56 shreds, yielded $41.8\%$ of accuracy. As mentioned by the authors, scalability for larger instances (i.e., with more shreds) is still an issue firstly due to the OCR working overhead and its accuracy degradation. Additionally, for better accuracy, the number and size of candidate sequences have to be increased, which compromises the run-time performance (it performed approx. 16 times slower than \cite{paixao2018deep} for a 60-shred instance).

In the last years, deep learning started to be used in the context of jigsaw puzzle solving
with simulated-cut tiles. Le and Li \cite{le2018jigsawnet} applied convolutional neural networks (CNNs) to verify potential matching pieces in order to reduce the search space for the posterior optimization process. Paumard et al. \cite{paumard2018jigsaw} solved small (3$\times$3) 2D-tile puzzles following the seminal ideas introduced in \cite{doersch2015unsupervised,noroozi2016unsupervised}, in which CNNs are trained in a self-supervised way to predict the relative positions of patches cropped from a reference image. More related to our work, Sholomon et al. \cite{sholomon2016dnn} used a fully connected network to measure pairwise compatibility between 2D-tile. Boundary pixels of two tiles are fed to the network and the network's output, i.e., the predicted adjacency probability, is assigned as the pair compatibility. Although these methods can improve the results, they only considered a non-realistic scenario with simulated-cut pieces.

To the best of our knowledge, this work is the first to explore deep learning in a realistic scenario with multiple shredded documents, having a preliminary investigation presented in \cite{paixao2018deep}. The use of deep models aims to improve robustness in real shredding context, where the damage to the shreds' borders prevents the use of similarity evaluation at pixel-level or of stroke continuity analysis. Additionally, our approach is able to cope with more heterogeneous content because the fitting of patterns is learned in a self-supervised fashion from large-scale data without segmenting symbols, as detailed in the next section.

\begin{figure*}[htpb!]
	\centering
	\includegraphics[width=\textwidth]{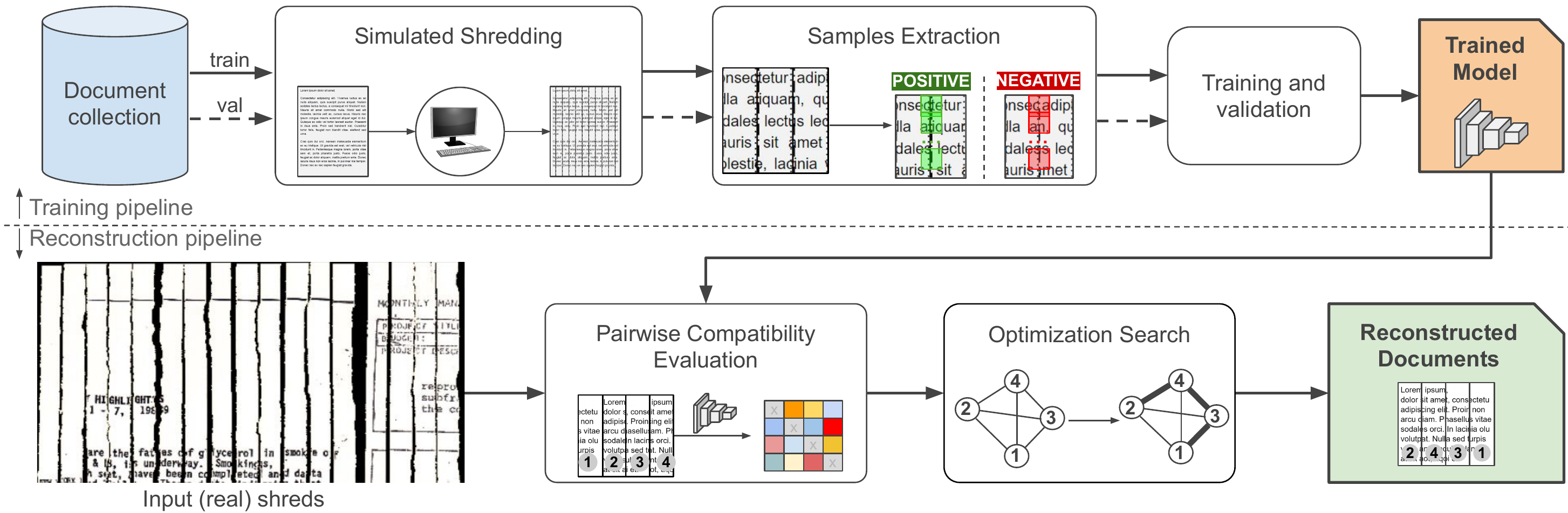}
	\caption{Overview of the proposed system. The training pipeline (top flow) comprises the generation of training data from simulated-shredded documents followed by the training itself, where the best-epoch model is chosen through validation. The reconstruction pipeline (bottom flow) represents the system in operation. The input is a set of real shreds (from one or more documents) and the output is a permutation of shreds (reconstructed documents). The trained model is used for shred's pairwise compatibility evaluation, and the resulting values (arranged as a matrix) are the inputs for a graph-based optimization search whose solution represents the intended shreds' permutation.}
	\label{fig:overview}
\end{figure*}
\section{Deep Learning-based Reconstruction}
\label{sec:proposed}

The proposed reconstruction system is essentially divided into training (off-line) and reconstruction (on-line) pipelines (\fig{fig:overview}).
The training (top flow) aims to produce a model capable of quantifying the compatibility between shreds based solely on the content around the cutting sections of digitally-cut documents.
Small samples (given the whole document) extracted from these documents are the patterns to be learned. This local approach follows the intuition behind the manual
reconstruction, where humans analyze the fitting of shreds based on local matching of
fragmented patterns (mainly at text line level).
These samples should be categorized as positive if they are likely to appear on real documents, or negative otherwise. In practice, positive samples are cropped from pairs of adjacent shreds, and negative from non-adjacent pairs. The learning process is said to be self-supervised because the adjacency relationship is automatically inferred in simulated shredding. After sampling, the deep model -- a fully convolutional neural network (FCNN) -- is trained as a classification model to distinguish between positive and negative samples, being the best model parameters determined through a validation process also using simulated data. 

For reconstruction (bottom flow), the system has a mild assumption: the shreds of the documents to be reconstructed are already individualized in digital format, i.e., the documents were previously shredded, scanned, and their shreds were segmented. The best model obtained in the training stage is used for pairwise compatibility evaluation of the shreds. The resulting values, arranged as a square matrix, are the input for a graph-based optimization procedure that estimates a shreds' permutation representing the final reconstruction. The training and reconstruction pipelines are presented in the following subsections alongside a more in-depth description of the proposed system.

\begin{figure}[t]
	\centering
	\includegraphics[width=0.75\textwidth]{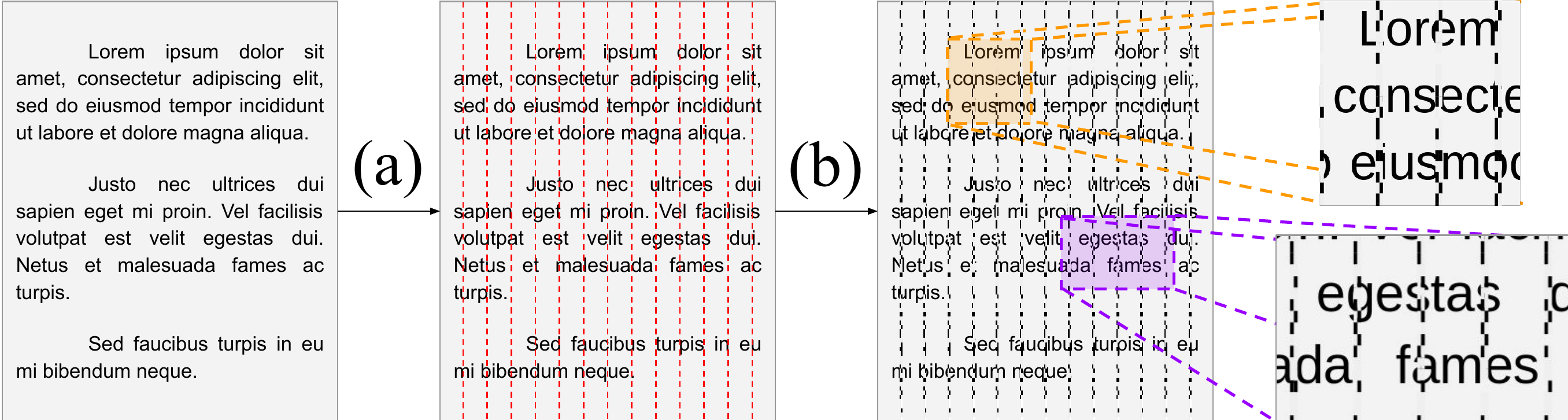}
	\caption{Simulated shredding. The document is digitally shredded into (a) longitudinal cuts (i.e., strip-cut shredding) and random noise is added (b) to the shreds' borders to roughly simulate the damage caused by paper shredders.}
	\label{fig:simulated_shredding}
\end{figure}

\subsection{Learning from Simulated-shredded Documents}
\label{subsec:training}

The training pipeline aims to produce a model capable of quantifying pairwise compatibility, which means the probability of two shreds to be adjacent in a certain order (order matters given the nature of the reconstruction problem). The input can be any collection of digital documents from which all the training data is extracted through simulated shredding. This is particularly beneficial since there is a lack of publicly available datasets containing real-world textual shredded documents, and the generation of such a kind of dataset is tedious, error-prone, and highly demanding because it requires printing, submitting the documents to a paper shredder, manually organizing and scanning the shreds, and, finally, post-processing them. The generation of training data and the training process itself are discussed in the next sections.

\subsubsection{Simulated Shredding}

This process consists of longitudinally slicing digital documents into $30$ rectangular regions with the same dimensions, wherein the height of the regions is equal to that of the input image, and the number of regions is an approximation of the number of shreds produced by regular shredders for A4 paper sheets. Text documents in most available public collections are binary or almost binary (e.g., the two collections in \sect{sec:datasets}). This motivated us to adopt a binary representation of the input documents -- resulting from applying Sauvola's method \cite{sauvola2000adaptive} -- before the simulated shredding.

The simulated shreds, however, present clean edges, which is very unlikely in real-world mechanical shredding.
To cope with that, the original content of the two rightmost and leftmost pixel columns of the shreds is replaced by a black-and-white pattern drawn from the uniform binary distribution $U(0, 1)$.
An overview of the process is depicted in \fig{fig:simulated_shredding}.

\subsubsection{Sample Extraction}
\label{sec:samples}

The input of this step is a document-wise set of digital shreds, and the output is a set of samples to be used in the training of the deep learning model. Given an input document, samples are extracted by pairing shreds and cropping small regions around the touching borders: positive samples come from adjacent shreds (respecting the shreds' ground-truth order) and negatives from non-adjacent shreds (or adjacent shreds in swapped order). As the shreds are automatically obtained, the samples can be self-annotated since the correct sequence of shreds in the original document is known.

Positive and negative training samples were extracted following the same procedure: given a pair of shreds, a sample is a rectangular region of $32\times32$ pixels ($32$ rows of the $16$ rightmost pixels of the left shred and $32$ rows of the $16$ leftmost pixels of the right shred). Such dimensions correspond to the minimum even-valued input size that the adopted network architecture (described in the next section) can handle.

The shreds were sampled every two pixels along the vertical axis, and a limit of $1{,}000$ positive samples per document was fixed. To produce balanced sets, the number of negative samples is limited to the number of positive samples collected in the same document. 
It is important to mention that the document datasets are available in binary format, as further discussed in \sect{sec:datasets}.
In this context, we define the level of information of a sample as the percentage of its text (assumed as black) pixels.
For effective training, samples with an information level less than a threshold $\rho_{black}$ are discarded due to the class ambiguity of such cases. This threshold was empirically set to $0.2$ based on visual inspection of a few samples: lower than this value, samples usually look like scanning noise.

Before extraction, the pairs of shreds are firstly shuffled to ensure sampling in different regions of the document since the number of samples per document is limited. Note that the extraction procedure is applied to each fragmented document obtained with simulated shredding, one document at a time, resulting in balanced sets of positive and negative samples.

\subsubsection{Model Training}
\label{sec:training}

At this point, two balanced sets (positive and negative) of shreds with $32\times32$ pixels are available for training the deep learning model. The SqueezeNet \cite{iandola2016squeezenet} architecture pre-trained on $227\times227\times3$ (RGB) images\footnote{The $224\times224\times3$ size reported in \cite{iandola2016squeezenet} seems a typo since $227\times227\times3$ is the size used in the official implementation (\url{https://github.com/forresti/SqueezeNet}).} of ImageNet was chosen because it has been shown to be efficient for the classification task, i.e., it can achieve good performance with considerable few parameters, and due to its fully convolutional structure, which is particularly interesting during the inference time, as further discussed in \sect{subsec:inference}. More specifically, the vanilla (i.e. no bypass connections) SqueezeNet v1.1 implementation was adopted, which is a modification of the original SqueezeNet with similar accuracy in the classification task, however, with 2.4 times less computation effort\footnote{\url{https://github.com/forresti/SqueezeNet/tree/master/SqueezeNet_v1.1}.}.

Since SqueezeNet is fully convolutional, it can be fed with images whose dimensions are different from the original input size. Therefore, training with $32\times32$ samples does not require any further architectural modifications. 
To leverage the ImageNet's pre-training, the binary samples were replicated to the three channels of the network instead of reducing the network's input to a single channel. The number of filters in the last convolutional layer was reduced from $1{,}000$ (ImageNet's number of classes) to two filters in order to match the positive and negative classes, and the weights for this layer were initialized under a zero-mean Gaussian distribution with a standard deviation of $0.01$, as done in the original SqueezeNet implementation.

From the entire database, $90\%$ of the documents (random selection) were designated for training, and $10\%$ for the validation of the model.
Therefore, samples of the same document are used exclusively either to train or validate the model. With the architecture properly adjusted for the problem and the weights initialized, the training can begin. The model was trained during $10$ epochs in mini-batches of $256$ images using the Adam optimizer with default settings \cite{kingma2014adam} and the categorical cross-entropy loss. The classification accuracy was measured on the validation set at the end of each epoch, and the epoch that yielded the highest accuracy was chosen to determine the ``best'' model, i.e., the model deployed for compatibility evaluation.

\subsection{Reconstruction of Mixed Shredded Documents}
\label{subsec:inference}

The system described in \fig{fig:overview} (bottom pipeline) assumes that the documents were previously fragmented by a paper shredder and that the resulting shreds were scanned and separated in image files at a disk (the semi-automatic segmentation process adopted by our group is detailed in \cite{paixaotifs2019}).
After loading data, the shreds are also binarized with Sauvola's algorithm \cite{sauvola2000adaptive} since the model was trained with binary samples. Subsequently, as recommended in \cite{prandtstetter2008}, the blank shreds (i.e., those without black pixels) are discarded since they increase processing overhead without providing relevant information for the forensic examiners. Then, the trained neural model is applied for pairwise compatibility evaluation of the remaining (non-blank) shreds.
These compatibility values (arranged as a matrix) are the inputs for the optimization search of the reconstruction problem solution: the permutation of shreds that (ideally) reassembles the original documents. Both compatibility evaluation and optimization search are discussed in the next subsections.

\begin{figure}[t]
	\centering
	\subfloat[]{\includegraphics[scale=0.6]{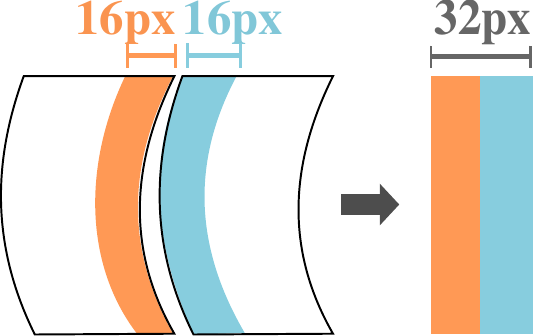} \label{subfig:rect_a}} \hfil
	\subfloat[]{\includegraphics[scale=0.6]{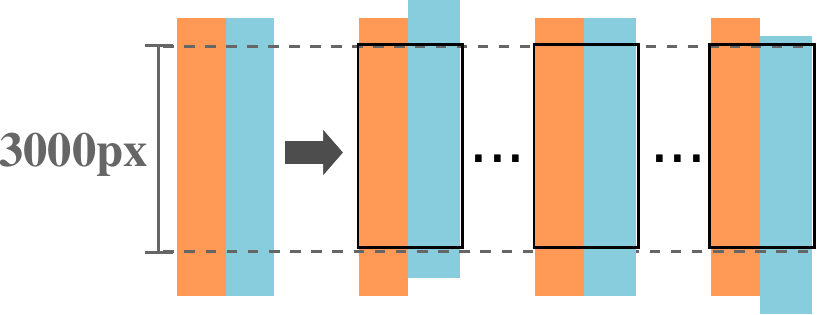} \label{subfig:rect_b}}%
	\caption{Association image extraction: (a) rectified interest regions; (b) network visual field definition by cropping and sliding operations.}
	\label{fig:rect}
\end{figure}

\subsubsection{Pairwise Compatibility Evaluation}
\label{sec:pairwise_comp_eval}

Given a set of $N_{shreds}$ non-blank shreds $\mathcal{S} = \{s_1, s_2, \cdots, s_{N_{shreds}}\}$, resulting from mixing the shredded documents to be reconstructed, an ordered pair $p_{ij} = (s_i, s_j) \in \mathcal{S}^2$, $i \neq j$, denotes $s_i$ placed to the left of $s_j$.
The goal of this stage is to estimate a compatibility value for every $p_{ij}, i, j = 1,  \ldots, N_{shreds}, i\neq j$, which can be arranged in a square matrix $\mathbf{C}_{N_{shreds} \times N_{shreds}}$ where each entry $(i, j$) matches the compatibility for $p_{ij}$. In other words, $\mathbf{C}_{ij}$ quantifies how likely $s_j$ is the right neighbor of $s_i$ in the original document. The estimation of $\mathbf{C}_{ij}$ is focused on regions around the edges of $s_i$ and $s_j$ (see \fig{subfig:rect_a}). The $16$ rightmost pixels of each row of $s_i$ are joined (at left) with the $16$ leftmost pixels of $s_j$, giving rise to a $H\times32$ rectified image, where $H$ is the minimum height of both shreds. The rectified image carries the information to be evaluated by the trained model. To account for vertical misalignment, different images are derived from the rectified image by vertically shifting its right part (blue area) $s$ units in the range $[-v_{shift}, v_{shift}]$. Let each of these images to be denoted by $I_{s}$, the subscript $s$ indicating the vertical shift. By default, $v_{shift}$ is set to $10$, thus $21$ (i.e., $2v_{shift} + 1$) different images should be evaluated. Only the $3{,}000$ center rows are considered in computation, as illustrated in \fig{subfig:rect_b}.

\begin{figure}[t]
	\centering
	\includegraphics[width=0.55\textwidth]{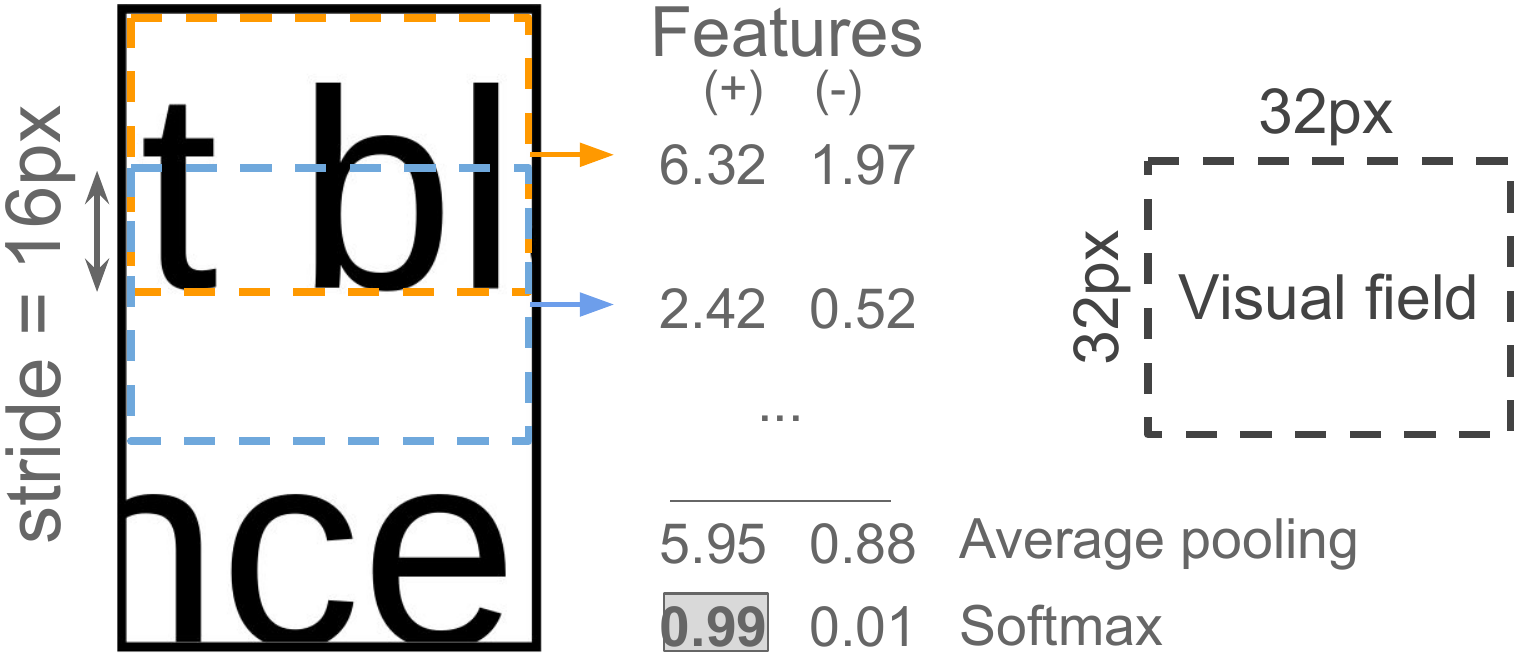}
	\caption{Compatibility computation as a sliding window operation. Since the model was originally trained on $32\times32$ images, applying it to $3000\times32$ images is equivalent to sliding vertically the $32\times32$-size model with implicit stride of $16$. In the example, the compatibility candidate value is $0.99$, the positive softmax probability.}
	\label{fig:slide_window}
\end{figure}

For faster inference, the derived images are bundled in a batch of size $21$ and processed by the deployed neural network.
Since the SqueezeNet architecture is fully convolutional and was trained with images of $32\times32$ pixels, the inference on a $3000\times32$ image is equivalent to sliding vertically the $32\times32$-size trained network across the input image with an implicit stride of $16$, as illustrated in \fig{fig:slide_window}. Note that inference on $32\times32$ pixels produces a pair of feature values (positive and negative). When applied to a $3000\times32$ image, an inference produces a $187\times2$ feature map ($187 = \lfloor \frac{3000}{16} \rfloor$). After global average pooling, the map is reduced to a pair of positive/negative logits from which probabilities are obtained via softmax. The compatibility $\mathbf{C}_{ij}$ is then set to the highest positive probability in a total of $21$ values. More formally,

\begin{equation}
\label{eq:c_ij}
\mathbf{C}_{ij} = \operatorname{max}_{s \in [-v_{shift}, v_{shift}]}{\sigma^{+}(\mathbf{y}(I_s))},
\end{equation}
where $\mathbf{y}(I)$ represents the network's logits output given the image $I$, and $\sigma^+(\mathbf{y})$
the positive probability computed by the softmax function on $\mathbf{y}$.

\subsubsection{Optimization Search}
\label{sec:solving}

The solution of the reconstruction problem is represented by a sequence $s_{\pi_1} s_{\pi_2} \ldots s_{\pi_{N_{shreds}}}$ of $\mathcal{S}$, where $\pi = \pi_1\pi_2\ldots\pi_{N_{shreds}}$ is a permutation of $\{1, 2, \allowbreak \ldots, N_{shreds}\}$. The goal of this stage is to estimate a permutation $\pi$ from the previously computed compatibilities. To accomplish this, it was adopted the graph-based optimization model proposed by our group in \cite{paixaotifs2019}, which is closely related to \cite{morandell2008,prandtstetter2008}. The central idea is that the desired solution (permutation) is given by solving the Minimum-cost Hamiltonian Path Problem (MCHPP) of a weighted directed graph that represents the shreds and association costs (incompatibility) among them.

Given that MCHPP is a minimization problem, a distance (cost) matrix $\mathbf{D}_{N_{shreds} \times N_{shreds}}$ is first derived from the compatibility matrix $\mathbf{C}$ by setting $\mathbf{D}_{ij} = \operatorname{max}(\mathbf{C}) - \mathbf{C}_{ij}$ for $i,j = 1, 2, \ldots, N_{shreds}, i \neq j$, and $+\infty$ for the diagonal elements of $\mathbf{D}$. The distance matrix can be viewed as a complete directed weighted graph $G = (V=\{v_1, \ldots, v_N\}, A, w)$, where a vertex $v_i \in V$ maps to a shred $s_i \in \mathcal{S}$, $A \in V^2$ is the set of arcs, and $w : A \rightarrow \mathbb{R}$ is the weight function defined such that $w((v_i,v_j)) = \mathbf{D}_{ij}$. As shown in \cite{paixaotifs2019, paixao2018deep}, the solving formulation consists in first converting the MHCPP problem into an Asymmetric Traveling Salesman Problem (ATSP) by adding a ``dummy'' vertex $v'$ and connecting it to all other vertices with zero weight arcs. ATSP is solved exactly as proposed in \cite{paixaotifs2019}. In summary, ATSP is reduced to the (Symmetric) Traveling Salesman Problem (TSP) by the two-node transformation proposed by Jonker and Volgenant \cite{jonker1983}. The solution for TSP is provided by an exact solver (Concorde software \cite{applegate2003} configured with QSOpt\footnote{\url{http://www.dii.uchile.cl/~daespino/QSoptExact\_doc/main.html.}}).
\section{Experimental Methodology}
\label{sec:experimental}

The general purpose of the experiments is to evaluate the reconstruction accuracy by mixing different quantities of single-page shredded documents (hereafter referred to as documents for simplicity) following an incremental strategy. Besides the two evaluation datasets used in \cite{paixaotifs2019} and \cite{paixao2018deep}, a new collection with 100 documents was assembled specifically for this investigation.

The experiments were divided into three main parts. First, the proposed method was evaluated in its default configuration.
Then, an ablation study was conducted to assess the sensibility of our method concerning three key parameters.
The last part is a comparative evaluation with state-of-the-art methods available in the literature. The following sections describe,
respectively, the datasets and the performance metric used for quality assessment, the conducted experiments, and, finally,
the computational platform on which the experiments were carried out.

\subsection{Training Datasets}
\label{sec:datasets}

As stated in \sect{subsec:training}, the training of the model for compatibility evaluation should take, as input, any document collection. In fact, different training datasets should be provided to enable cross-database evaluation. In this work, two collections of scanned documents were used (one at a time) to extract training samples: \DISRI and \DCDIP.

\subsubsection{\DISRI}
In this paper, \DISRI refers to a subset of the ISRI-Tk OCR collection \cite{nartker2005} used in \cite{paixao2018deep},
which includes $800$ binary documents (originally scanned at $300$ dpi) labeled as \textit{reports}, \textit{business letters}, or
\textit{legal documents}. The structure of these documents has a high degree of similarity, generally focusing on running text at the expense of graphical elements (i.e., pictures, tables, graphs).

\subsubsection{\DCDIP}

This dataset comprises $100$ documents from the RVL-CDIP collection \cite{harley2015icdar}, of which there are $10$ documents from each of the following classes: \textit{form}, \textit{email}, \textit{handwritten}, \textit{news article}, \textit{budget}, \textit{invoice}, \textit{questionnaire}, \textit{resume}, and \textit{memo}.
In summary, this dataset has a more diverse collection of documents.
The documents were chosen arbitrarily, except for the restriction that they should present textual content at some level.
Since the RVL-CDIP is a subset of the IIT-CDIP Test Collection 1.0 \cite{lewis2006building} but in lower resolution, we decided to use the corresponding $300$ dpi images from the original IIT-CDIP dataset (the resolution matches that of the evaluation datasets).

\begin{figure}[t]
	\centering
	\subfloat[]{\includegraphics[width=0.2\textwidth]{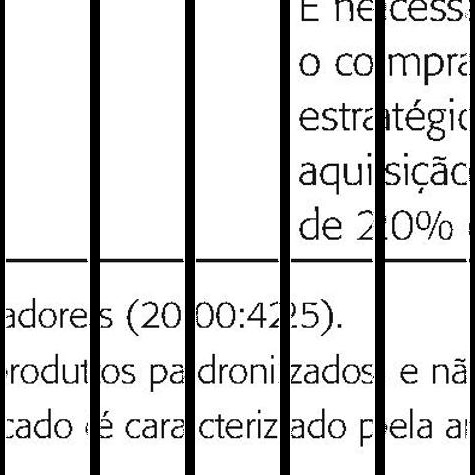}%
		\label{fig:datasets:D1}}
	\hspace{2em}
	\subfloat[]{\includegraphics[width=0.2\textwidth]{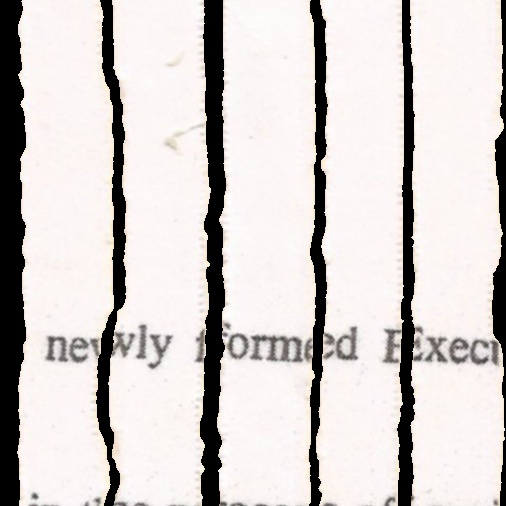}%
		\label{fig:datasets:D2}}
	\hspace{2em}
	\subfloat[]{\includegraphics[width=0.2\textwidth]{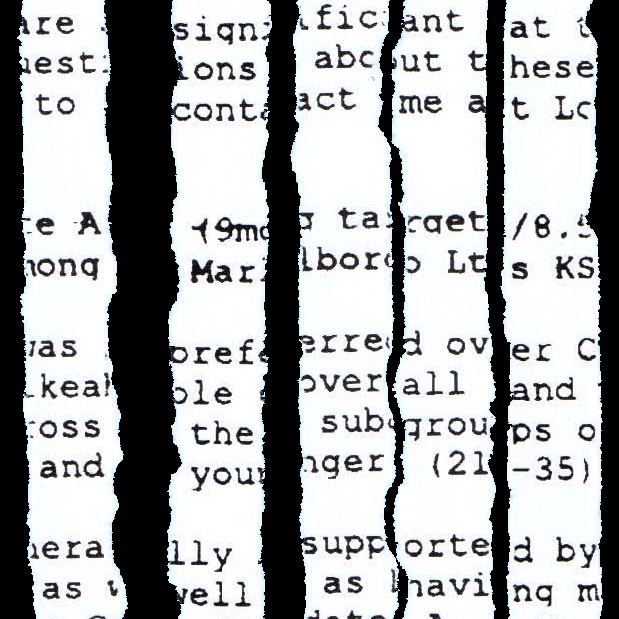}%
		\label{fig:datasets:D3}}
	\caption{Samples (cropped view) of the shredded (denoted by the ``S-'' prefix) datasets: (a) \DSMARQUES; (b) \DSISRI; (c) \DSCDIP.}
	\label{fig:datasets}
\end{figure}

\subsection{Evaluation Datasets}

Three datasets were used to evaluate the methods: \DSMARQUES, \DSISRI, and \DSCDIP (the last is a contribution of this work).
The ``S-'' prefix stands for mechanically ``shredded'' and was used to differentiate from the training datasets, which comprise the original (unshredded) documents.

\subsubsection{\DSMARQUES}
\label{sec:s-marques}
\DSMARQUES refers to the $60$ text documents in Portuguese of the strip-shredded dataset produced by Marques and Freitas \cite{marques2013}. To create this dataset, the authors collected $60$ paper documents (a digital backup is also available along with the dataset) and shredded them using a Cadence FRG712 strip-cut machine. The resulting shreds were scanned at $300$ dpi and then separated in JPEG files (one for each shred). Compared to the other datasets, as shown in \fig{fig:datasets}, \DSMARQUES's shreds have a more uniform shape, and are less damaged by the shredder's blades, i.e., they are less curved and their borders are less corrupted (smooth serrated effect).

\subsubsection{\DSISRI}
\label{sec:s-isri}

This dataset was produced in the previous work of our research group \cite{paixaotifs2019,paixao2018deep} from a set of $20$ business letters and legal reports of the ISRI-Tk OCR collection, the same set used in \cite{andalo2017} to assess the reconstruction of simulated-shredded documents.
The digital documents were printed onto A4 paper and subsequently submitted to a Leadership 7348 strip-cut paper shredder.
To expedite the acquisition process, the shreds were spliced onto a high-contrast paper, and, after scanning (at $300$ dpi), they were segmented and stored individually in JPEG files.
This process is more detailed in \cite{paixaotifs2019}.

\subsubsection{\DSCDIP}
\label{sec:s-cdip}

The \DSCDIP dataset, which is a particular contribution of this work, is the shredded version of the $100$ digital documents in \DCDIP. The same methodology to create \DSISRI was also adopted for this dataset. As illustrated in \fig{fig:datasets}, the shreds of \DSISRI and \DSCDIP depict a higher degree of vertical misalignment in view of \DSMARQUES, as well as more damage at their extremities.

\subsection{Multi-reconstruction Accuracy Metric}
\label{sec:metrics}

The reconstruction quality for strip-shredded documents is often measured as the proportion of matching shreds in the reconstruction solution \cite{prandtstetter2008,morandell2008,andalo2017,paixao2018deep,paixaotifs2019}, i.e.,
the number of pairs of adjacent shreds in the estimated solution that are also adjacent in the original document.
For multi-reconstruction, nevertheless, the solution includes shreds from different documents, and the order these documents appear is not relevant.
To account for this fact, we consider that, in a multi-reconstruction solution, the last shred of a document matches the first shred of any other document.

For a more formal definition, let $\mathcal{S}^{(d)} = \{s^{(d)}_i : i = 1, 2, \ldots, N^{(d)}_{shreds}\}$ denote the set of $N^{(d)}_{shreds}$ shreds (indices indicating the ground-truth order) of the $d$-th individual document among those to be reconstructed.
From the notation introduced in \sect{sec:pairwise_comp_eval}, it follows that $\mathcal{S} = \bigcup_{d}{\mathcal{S}^{(d)}}$ and $N_{shreds}=\sum_{d}{N^{(d)}_{shreds}}$. A solution for the reconstruction problem is given by a bijective mapping $R$ of the documents shreds' into positions, i.e., $R : \mathcal{S} \rightarrow \{1, 2, \ldots, N_{shreds}\}$.
For multi-reconstruction, a matching pair of shreds in a solution has the form $(s^{(d)}_{i}, s^{(d)}_{i+1})$ (intra-document) or $(s^{(d_1)}_{last}, s^{(d_2)}_1)$ (inter-documents), where $last = N^{(d_1)}_{shreds}$ and $d_1 \neq d_2$.
Therefore, following these two matching criteria, accuracy can be calculated as

\begin{equation}
\label{eq:accuracy}
accuracy=\frac{\sum_{i=1}^{N_{shreds} - 1} \operatorname{\bf 1}[R^{-1}(i) \text{ matches } R^{-1}(i + 1)]}{N_{shreds} - 1}
\end{equation}
where $\operatorname{\bf 1}[\cdot]$ is the 0-1 indicator function. Note that accuracy ranges in the interval $[0,1]$, where 0 implies a fully disordered reconstruction, and 1 is achieved only by a perfect reconstruction.

\subsection{Experiments}
In the preliminary work \cite{paixao2018deep}, the experiments were not cross-database since the documents of \DSISRI were reconstructed with a model trained on documents of the ISRI-OCR Tk collection. In practice, such experiments assume the availability of training data that share significant appearance and structural similarities with test data. For a more realistic scenario, the experiments in this paper followed a cross-database protocol in which testing on \DSCDIP (the dataset produced in this work) leverages the model trained on \DISRI, and testing on \DSISRI and \DSMARQUES uses the model trained on \DCDIP.

The evaluation is performed in an incremental way so that new shredded documents are gradually introduced to the reconstruction instance. For the sake of notation, let $k$ denote the number of mixed documents of a particular instance. The main purpose of the incremental approach is to evaluate whether the reconstruction accuracy degrades with the increasing of $k$. Due to the processing burden of this type of experiment, the ablation study evaluates incrementally $k = 1, 2, \ldots, 5$ documents at a time, while the other two experiments also include $k=10, 15, 20, \ldots, N_{docs}$, where $N_{docs}$ denotes the size of the current evaluation dataset ($60$, $20$, or $100$, as described in Section \ref{sec:datasets}). For each $k$ value, a set of $k$-size instances (i.e., $k$ mixed documents) should be sampled. Note that the size of the sample space varies significantly over $k$. For example, there are ${100 \choose 2} = 4{,}959$ possible ways of combining $2$ \DSCDIP's documents, whereas for $k=3$, this number rises to $161{,}700$. Instead of independently sampling combinations, the test instances are assembled in such a way that 
the $k$-size instances are obtained by adding a single document to each instance of size $k-1$. For a better description, let $(\mathcal{S}^{(d)})_{d = 1, 2, \ldots, N_{docs}}$ be a random sequence of the shredded documents of a particular dataset. The test instances for a particular $k$ consist of all possible $k$ consecutive documents, i.e., $(\mathcal{S}^{(1)}, \ldots, \mathcal{S}^{(k)})$, $(\mathcal{S}^{(2)}, \ldots, \mathcal{S}^{(k+1)})$, and so on, until $(\mathcal{S}^{(N_{docs} - k + 1)}, \ldots, \mathcal{S}^{(N_{docs})})$. Note that this yields overlapping of test instances for the same $k$, as well as across $k$ values.

In the first experiment, the incremental procedure was used to assess the robustness of the proposed method in its default parameters' configuration: $\rho_{black}=0.2$ and samples size of $32\times32$ (defined in \sect{sec:samples}), and $v_{shift}=10$ (defined in \sect{sec:pairwise_comp_eval}).

For the second experiment, an ablation study was carried out to evaluate the sensitivity of the system with respect to the aforementioned parameters, one at a time. The parameters' domain for $\rho_{black}$, sample size, and $v_{shift}$ were set to $\{0.1, 0.2, 0.3\},$ $\{32\times32, 32\times64, 64\times32, 64\times64\}$, and $\{0, 5, 10, 20\}$, respectively. The investigation of $\rho_{black}$ aims to verify the system's robustness with respect to the amount of information contained in the samples, which can vary for different font types and sizes. The desirable behavior is that the average accuracy holds for the widest possible range of $\rho_{black}$. The motivation behind the analysis of the sample sizes is confirming whether the locality assumption holds or not. Notice that training with $64$-width samples requires adjusting the input window to $3000\times64$ in the pairwise compatibility evaluation stage (\sect{sec:pairwise_comp_eval}). The analysis of $v_{shift}$ evaluates the need for (vertically) aligning the shreds at test time since no image processing was previously applied to this intent.

The final experiment aims at comparing our method (referred to as \textbf{Proposed}) against three relevant methods of literature. The first is referred to as Paix\~ao, a character shape-based method developed in previous work \cite{paixaotifs2019}.
The original implementation, intended for single-document reconstruction, uses caching of shape dissimilarities to improve time efficiency, which, on the other hand, compromises memory scalability for multi-document reconstruction. Therefore, reconstruction with this method was limited to $k=5$ documents. The second method is the one proposed by Liang and Li \cite{liang2019reassembling}, which is referred to as Liang. Due to time restrictions of the provided implementation\footnote{\url{https://github.com/xmlyqing00/DocReassembly}.}, the multi-reconstruction experiment was run only for the datasets \DSMARQUES and \DSISRI limited to $k=3$ documents. We adopted the parameters for the real-shredded instances 1 and 2 (total of 3) of the original work. For the matter of consistency, we configured the OCR software on which the Liang method relies to the Portuguese language when testing on \DSMARQUES. The last method, referred to as Marques \cite{marques2013}, relies on edge pixel dissimilarity for compatibility evaluation and was chosen due to its superior performance compared to other methods of literature, as can be seen in \cite{paixao2018deep,paixaotifs2019}. While Paix\~ao and \textbf{Proposed} share the same optimization formulation, Marques uses a simple greedy nearest-neighbor approach. Thus, for a fairer comparison, our system was also evaluated with the Marques' optimization model to emphasize the role of compatibility evaluation in producing accurate reconstructions. The modified method is referred to as Proposed-NN.

\subsection{Experimental Platform}

The experiments were carried out on two machines: (M1) an Amazon AWS instance with 8 vCPUs (2.3GHz), 60GB RAM, and a GPU NVIDIA Tesla V100 (16GB); (M2) an Intel Xeon E7-4850 v4 (2.10GHz) with 128 vCPUs, 252GB RAM. The ablation study was fully performed on M1. The methods Liang, Paix\~ao, and Marques, which do not require GPU processing, were conducted on M2. As Liang leverages OpenMP\footnote{\url{www.openmp.org}.} directives to improve efficiency, we used 240 threads (120 vCPUs) in the experiments. For \textbf{Proposed}/Proposed-NN, the compatibility evaluation in the experiments 1 and 3 was carried out on M1, while the optimization process was performed in M2 due to the large memory resources. The proposed system was implemented in Python with TensorFlow for training and inference, and with OpenCV for image processing. The code, pre-trained models, and datasets are publicly available at \url{https://github.com/thiagopx/deeprec-pr20}.
\section{Results and Discussion}
\label{sec:results}

\subsection{Proposed Method}

\begin{figure}[t]
	\centering
	\includegraphics[width=0.85\textwidth]{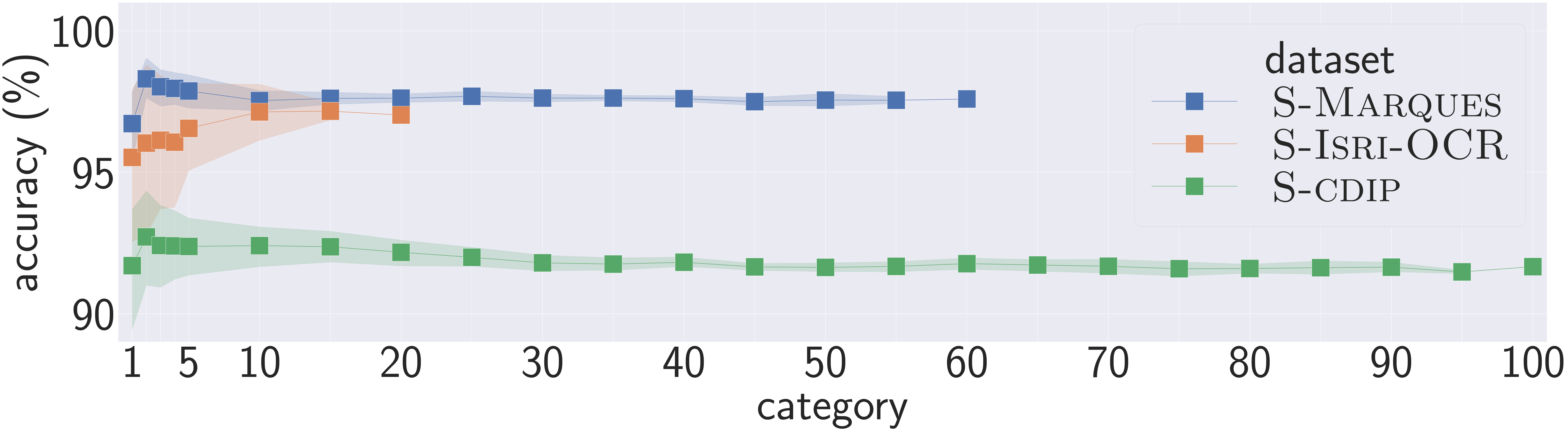}
	\caption{Multi-reconstruction accuracy across datasets. The square markers represent the mean values with respect to the documents instances, and the shadowed areas represent the $95\%$ confidence interval.}
	\label{fig:accuracy_proposed}
\end{figure}

\begin{figure*}[t]
	\centering
	\includegraphics[width=\textwidth]{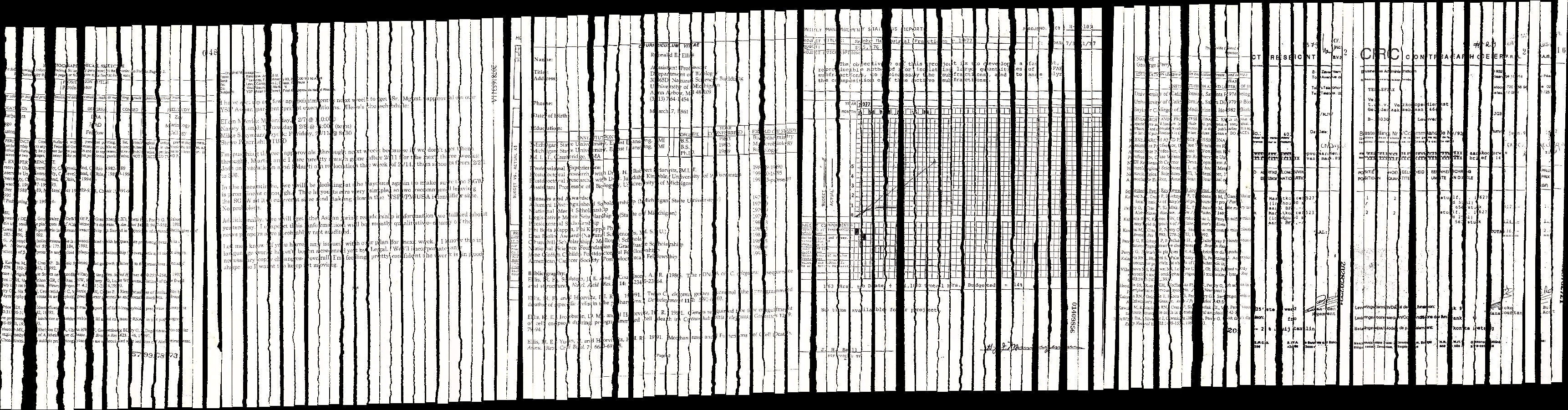} 
	\caption{Reconstruction of a test instance comprising shreds from $5$ documents (i.e., $k=5$) of \DSCDIP: the resulting accuracy was $82.93\%$. The shreds were placed side-by-side without any rotation correction. Each new inserted shred was vertically shifted according to the optimal $s$ value in \eq{eq:c_ij}. The full reconstruction ($k=100$) can be viewed at \url{https://htmlpreview.github.io/?https://github.com/thiagopx/docs/blob/master/results_s-cdip.html}.    
	}
	\label{fig:reconstruction_example}
\end{figure*}

\fig{fig:accuracy_proposed} shows the multi-reconstruction accuracy (mean and $95\%$ confidence interval) obtained with the proposed method (default parameters) for three evaluation datasets. Overall, the proposed method performed above $90\%$ for the three datasets, and, comparatively, \DSCDIP was verified, as expected, the most challenging test collection (an example of reconstruction is shown in \fig{fig:reconstruction_example}). The confidence interval tends to be wider as fewer documents are available, which is the case of \DSISRI. Furthermore, the accuracy tends to stabilize for large $k$, 
which means that the insertion of new documents into the reconstruction instance does not degrade accuracy, even though it increases considerably the complexity of the problem.

\begin{figure}[t]
	\centering
	\includegraphics[width=0.7\textwidth]{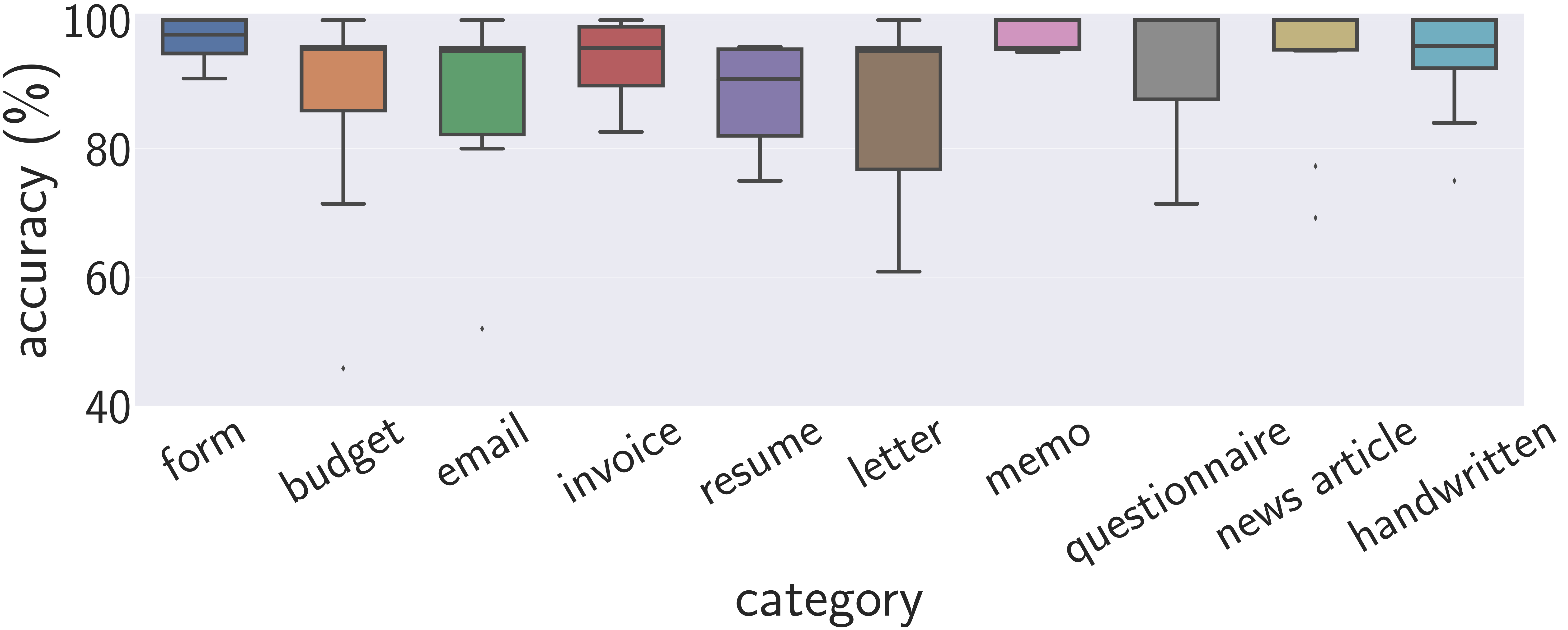}
	\caption{Reconstruction accuracy for \DSCDIP across categories ($k=1$).}
	\label{fig:accuracy_cdip_categories}
\end{figure}

Breaking down the performance on the \DSCDIP dataset, \fig{fig:accuracy_cdip_categories} shows accuracy boxplots for single-reconstruction ($k=1$) across the dataset categories. From the $100$ documents, $32$ were perfectly reconstructed, and only $5$ had accuracy lower than $70\%$. Remarkably, the reconstruction of handwritten documents achieved high accuracy ($8$ in $10$ were superior to $90\%$) although no handwritten document was used to train the compatibility evaluation model. For documents of the type form, all reconstructions achieved accuracy higher than $90\%$, being $5$ of them perfect. The results for the \textit{handwritten} and \textit{form} categories show that learning is not restricted to the symbolic level, and that lower-level features (e.g., strokes and horizontal lines) can also be learned by the model.

\begin{figure}[t]
	\centering
	\includegraphics[width=0.8\textwidth]{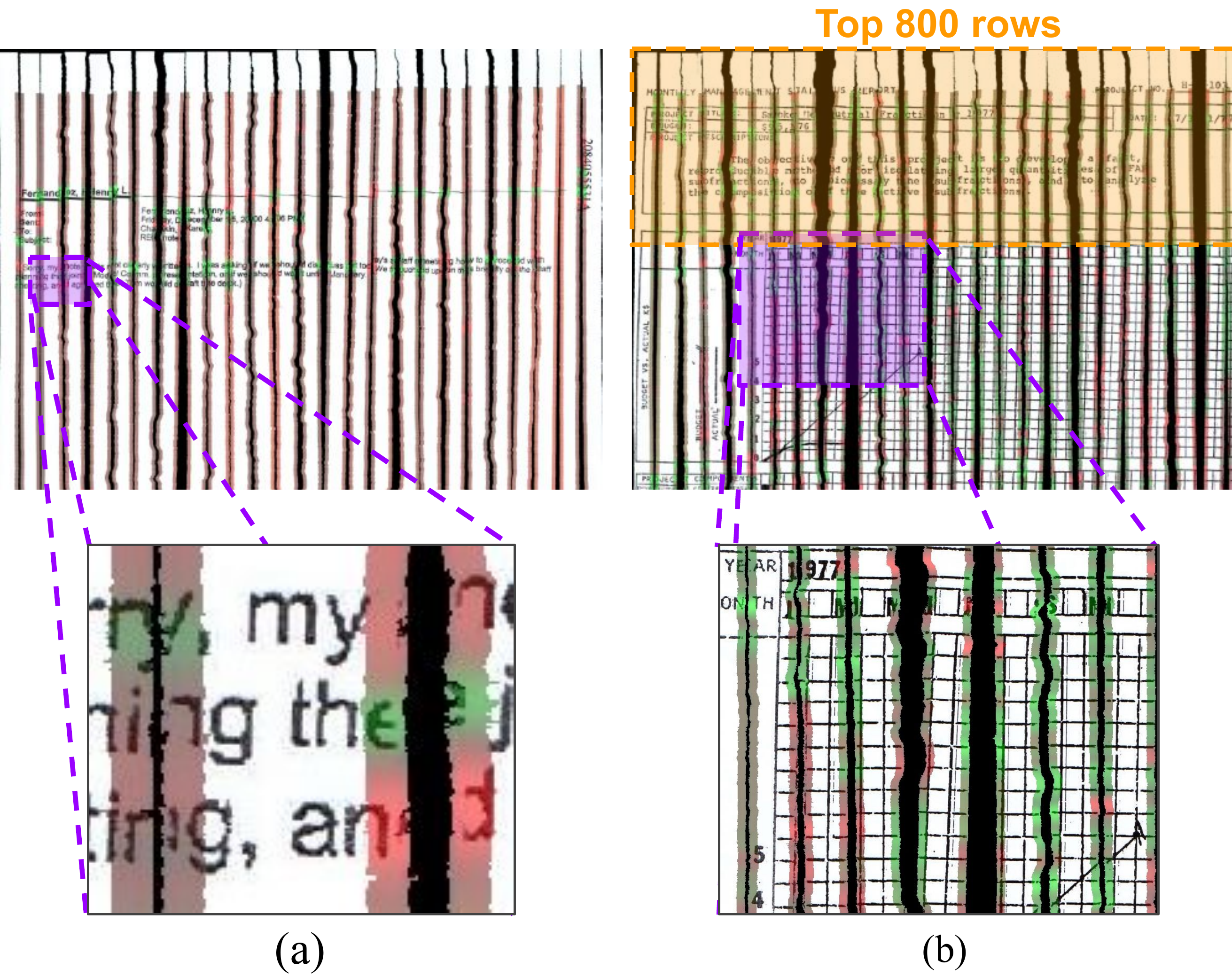}
	\caption{Challenging reconstruction instances (shreds are placed on the ground-truth order). Positive (green) and
		negative (red) activation maps for the adjacent shreds were superimposed onto the shreds. In (a), there is an email with large blank areas and corrupted characters for which the reconstruction accuracy was $52.00\%$.
		In (b), there is a budget document with a large grid pattern. The vertical lines of the grid induce negative activation (red) when they get close to the cut section (center of the network visual field). By cropping the top $800$ rows of the shreds (roughly indicated by the orange area), the accuracy rises from $45.83$ to $75.00\%$.
	}
	\label{fig:limitations}
\end{figure}

As seen in \fig{fig:accuracy_cdip_categories}, the \textit{letter} category has a larger variability in comparison to the others. In this category, there are three documents with very small fonts and whose shreds have degraded borders beyond the regular corruption found in most shreds. Although the accuracy for these three documents is low ($<75\%$), such values are not low enough to be considered outliers, which explains the elongated aspect of the letter's boxplot. The poor outlier performance, more evident in the \textit{budget} and \textit{email} categories, is mainly caused by three factors that may occur in combination or separately: (i) low quality of text symbols (i.e., low resolution, corrupted data), large flat areas (i.e., low amount of information), or (iii) large areas covered by patterns not learned by trained model.
These challenging factors are illustrated in \fig{fig:limitations}. The shreds were placed side-by-side in the ground-truth order and the activation maps from the SqueezeNet's last convolutional layer were adjusted and superimposed to the shreds boundary zones. Green areas represent high degree of compatibility, while red ones represent the opposite. Neutral zones are usually gray, indicating a balance between the positive and negative classes. Nonetheless, it can be noticed in \fig{fig:limitations}a reddish areas for neutral zones due to bias, or caused by noise (small black regions) in the highlighted areas
close to the borders.

In the first case (\fig{fig:limitations}a), an email document with large blank areas and corrupted characters was reconstructed with $52.00\%$ of accuracy. Due to the low amount of information, the compatibility evaluation and, as a consequence, the reconstruction accuracy is more sensitive to corrupted data. The second document (\fig{fig:limitations}b) is a budget with a large area covered by a grid pattern, and for which the obtained accuracy was $45.83\%$. Unlike the horizontal lines, which are captured by the model, the vertical lines lead to erroneous evaluations by the model.
This is justified by the scarcity of such patterns in the training set, which comprises images from \DISRI. By restricting the shreds to their first $800$ rows (orange highlighted region in \fig{fig:limitations}), the reconstruction accuracy increases to $75.00\%$. 
Although the aforementioned cases yielded low-accuracy reconstructions, it does not mean that the same behavior will invariably be observed for documents with similar layout/features.
The reconstruction quality also depends on where the cuts take place. In \DSCDIP, 
for example, there are an email and a budget document visually similar to those in \fig{fig:limitations} for which the accuracy was $80$ and $84\%$, respectively.

\begin{figure}[t]
	\centering
	\includegraphics[width=0.8\textwidth]{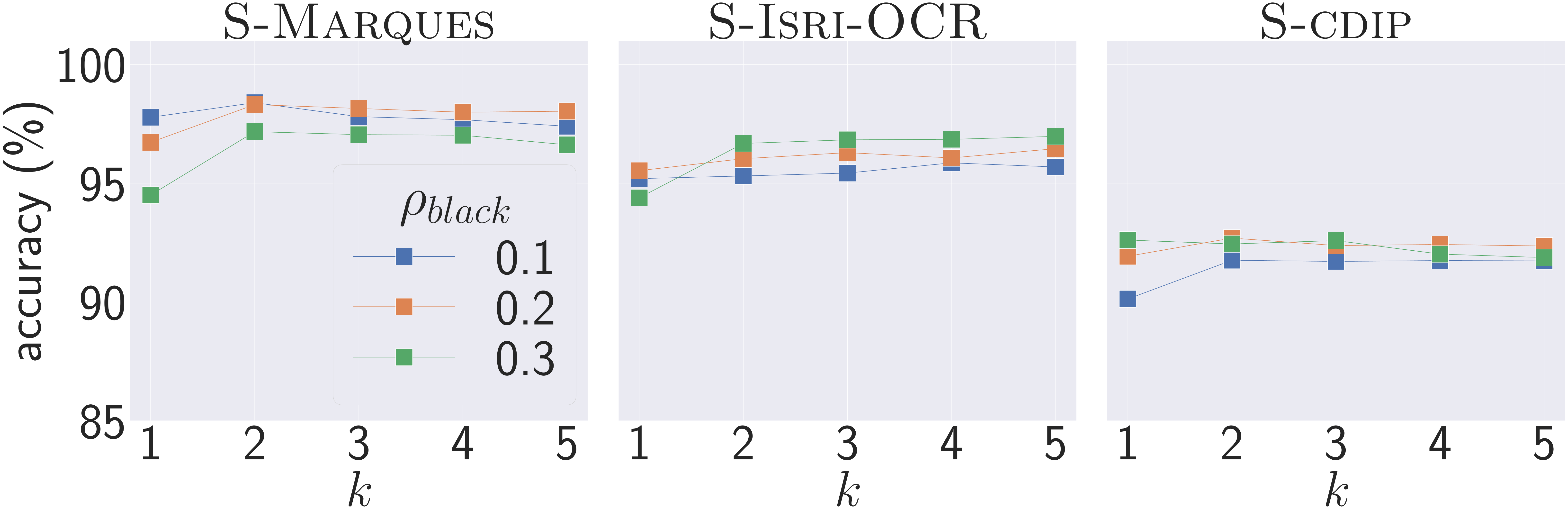} 
	\caption{Investigation of the accuracy sensitivity with respect to the parameter $\rho_{black}$.}
	\label{fig:neutral_thresh}
\end{figure}

\begin{figure}[t]
	\centering
	\includegraphics[width=0.8\textwidth]{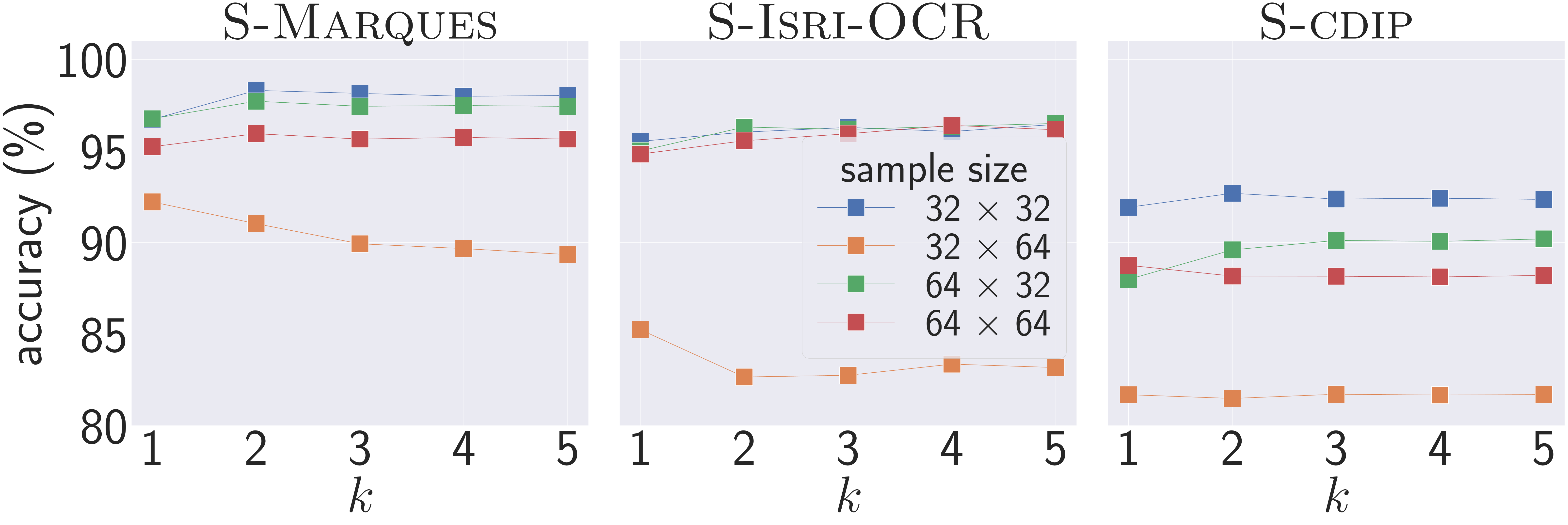}
	\caption{Investigation of the accuracy sensitivity with respect to the size of training samples.}
	\label{fig:input_size}
\end{figure}

\begin{figure}[t]
	\centering
	\subfloat[]{\includegraphics[width=0.2\textwidth]{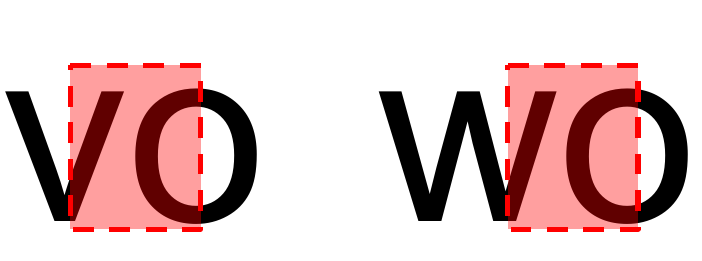}%
		\label{fig:ambiguity_a}}
	\hspace{2em}
	\subfloat[]{\includegraphics[width=0.2\textwidth]{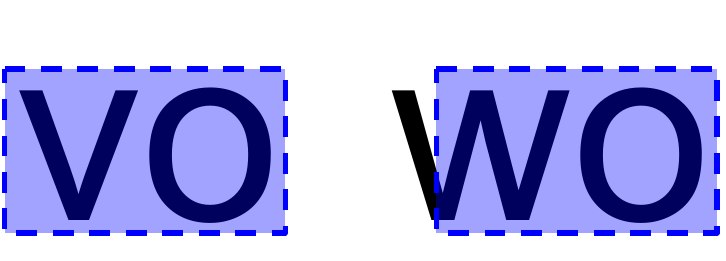}%
		\label{fig:ambiguity_b}}

	\caption{Visual ambiguity between ``wo'' and ``vo'' is illustrated in (a) for a $32\times32$ input window highlighted in red. Such ambiguity is not seen (b) after increasing the width of the input window (highlighted in blue).}
	\label{fig:ambiguity}
\end{figure}

\subsection{Ablation Study}

\begin{figure}[t]
	\centering
	\includegraphics[width=0.8\textwidth]{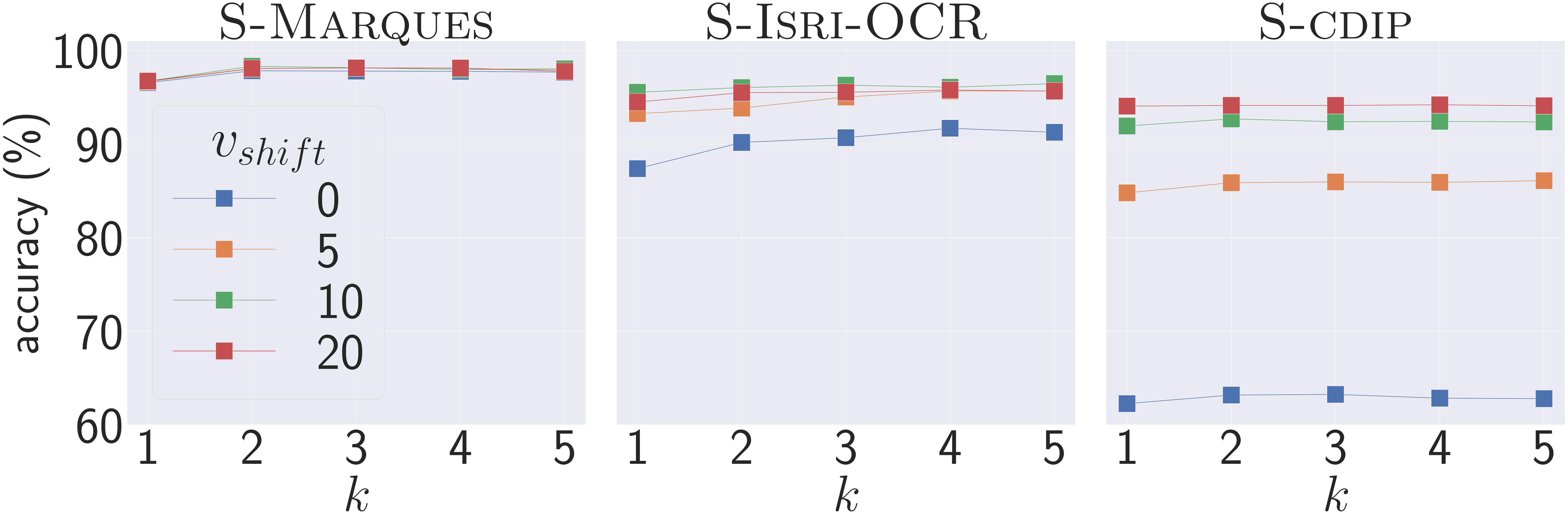} 
	\caption{Investigation of the accuracy sensitivity with respect to the parameter $v_{shift}$.}
	\label{fig:vshift}
\end{figure}

The results for the three investigated parameters are summarized in Figures \ref{fig:neutral_thresh}, \ref{fig:input_size}, and \ref{fig:vshift}. \fig{fig:neutral_thresh} shows the accuracy sensitivity with respect to the parameter $\rho_{black}$. Ideally, the system is expected to be robust to changes in this parameter. From the results, it can be observed a wider variation range for $k=1$. The performance difference becomes less noticeable as $k$ increases, which represents a more realistic scenario for the reconstruction application.

\fig{fig:input_size} shows the impact of the size of training samples on the final reconstruction performance. In general, the system generalizes better across the datasets for samples with reduced width, i.e., $32\times32$ and $64\times32$. By keeping the samples narrower, visual ambiguity (illustrated in \fig{fig:ambiguity}) can be explored in compatibility evaluation of scarce/unseen patterns in the training data. For instance, the model can perceive a ``wo'' association as valid (as in ``\textbf{wo}rld'') if samples with ``vo'' (as in ``\textbf{vo}xel'', ``\textbf{vo}lume'', and ``reser\textbf{vo}ir'') were observed during the training. The results for $64\times64$ samples were competitive in terms of accuracy on the \DSISRI, where the documents have primarily textual content. However, the performance dropped significantly for documents with higher density of graphical elements (e.g., forms and budgets) present in \DSMARQUES and \DSCDIP.

Finally, \fig{fig:vshift} shows the influence of the vertical shift range parameter ($v_{shift}$) on the reconstruction performance. In practice, no sensitivity to this parameter was observed for \DSMARQUES since the shreds for this collection are (practically) vertically aligned, as exemplified in \fig{fig:datasets:D1}. In contrast, the results on the \DSISRI and \DSCDIP datasets, which better depict real-world conditions, show the relevance of properly treating the misalignment between shreds. The misalignment degree is higher for \DSCDIP, which explains the consistent accuracy improvement with the increase of $v_{shift}$.

\subsection{Comparative Evaluation}
\begin{figure*}[ht]
	\centering
	\includegraphics[width=\textwidth]{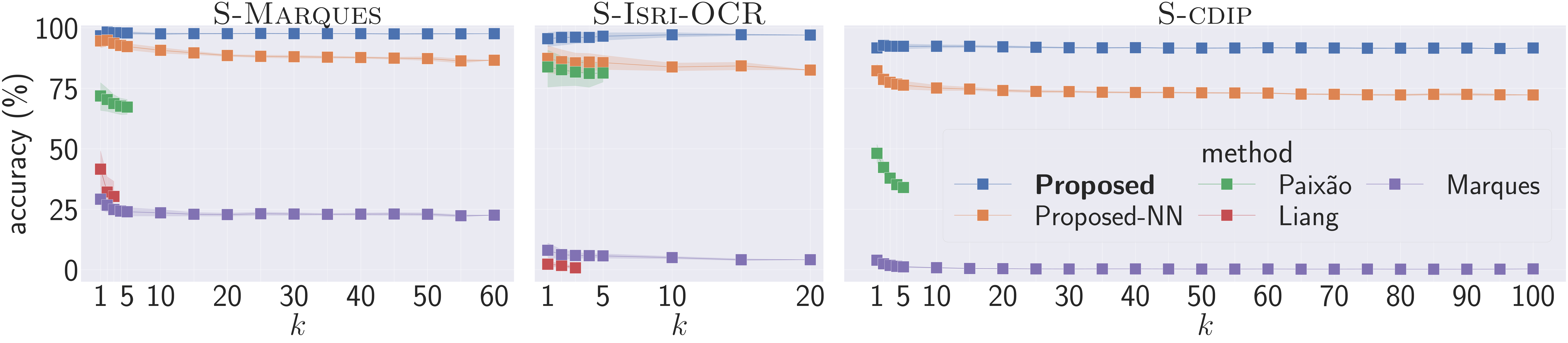}
	\caption{Comparative accuracy performance. The proposed approach achieved the highest accuracy for all three test sets.
	}
	\label{fig:comparison}
\end{figure*}

\fig{fig:comparison} shows the comparative performance with the literature. The average accuracy of the proposed method using Concorde (\textbf{Proposed}) was consistently superior to the compared methods. Additionally, it demonstrated greater robustness, which is mainly evidenced by the stability of the accuracy curve with the increase of $k$.

Unlike the proposed method, the modified version (Proposed-NN) -- intended for comparison with Marques -- presented a decay in accuracy with the increase of $k$. Nevertheless, it greatly outperformed Marques, which also uses the same optimization approach, and Paix\~ao, which leverages Concorde. In fact, Marques struggles with black-white documents since it is based on color features. Moreover, it is very sensitive to the damage on the shreds' borders caused by the mechanical fragmenting process, and to the vertical misalignment of the shreds. Both issues are accentuated in the \DSISRI and \DSCDIP datasets, resulting in a significant drop in performance when compared to \DSMARQUES. It can also be observed that the accuracy of Paix\~ao degrades more sharply for \DSCDIP, which is explained by the large presence of pictorial elements (as depicted in \fig{fig:limitations}b), and also by a greater diversity of symbols in different font types, sizes, and styles (including handwritten characters). When mixing documents, such diversity becomes a critical factor since Paix\~ao assumes a fixed-size alphabet in which each symbol has a unique representative. For single-reconstruction ($k=1$), Liang was capable of reconstructing 7 pages of \DSMARQUES (in a total of 60) with $100\%$ of accuracy. These instances have a great concentration of text and no pictorial content. Nonetheless, the average accuracy considering all the 60 pages was under $50\%$ with a sharp decay as $k$ increases. The observed decay corroborates the scalability issue raised by the authors and mentioned in Section \ref{sec:related_work}. Like Marques, the accuracy was dramatically worse for \DSISRI than for \DSMARQUES. This is because Liang strongly relies on the OCR capability of recognizing full words on composition of shreds (visually similar to \fig{fig:reconstruction_example}), and such capability is substantially affected by geometric distortion between shreds.


\begin{figure}[t]
	\centering
	\subfloat[]{\includegraphics[height=80pt]{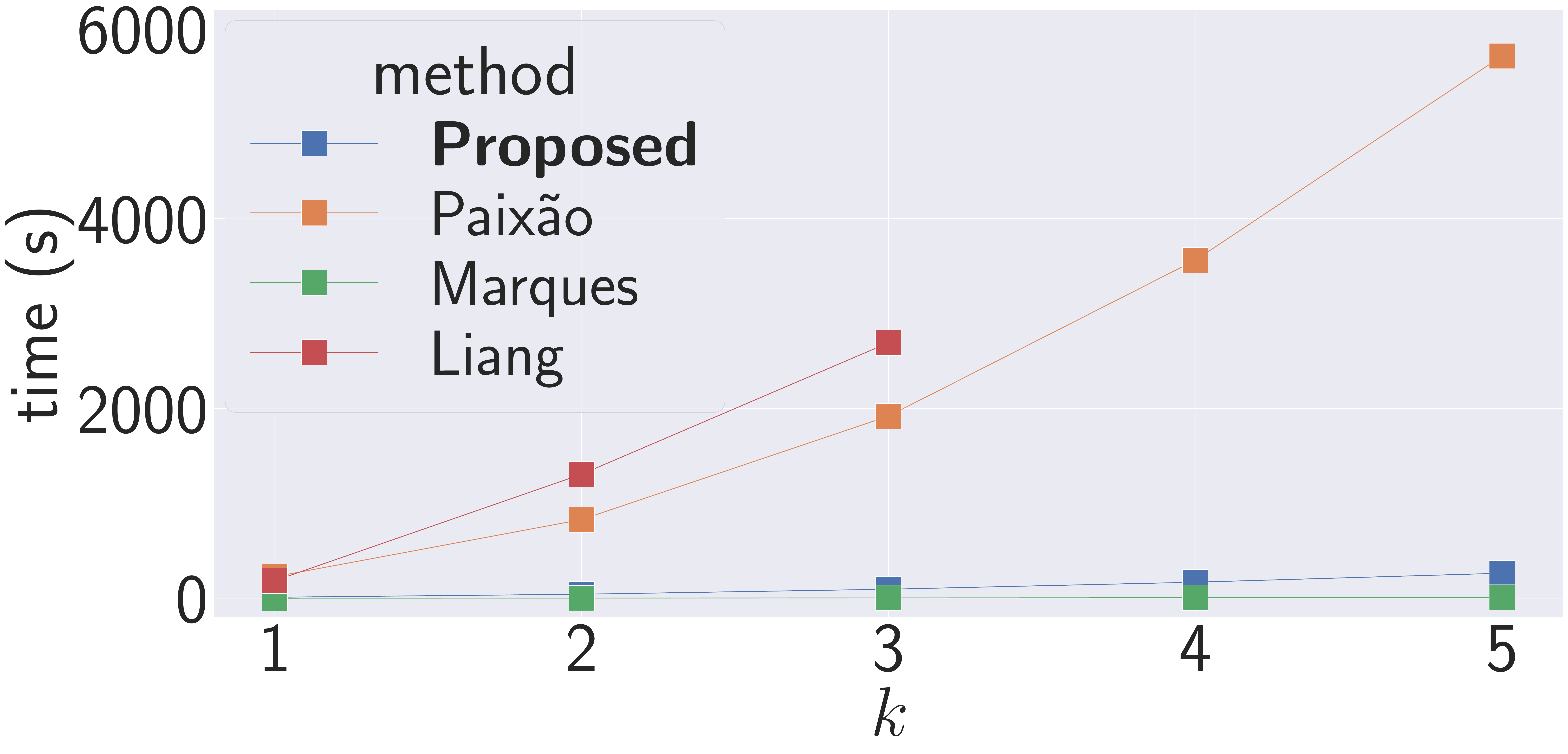}    \label{fig:time_1}}
	\hspace{1em}
	\subfloat[]{\includegraphics[width=80pt]{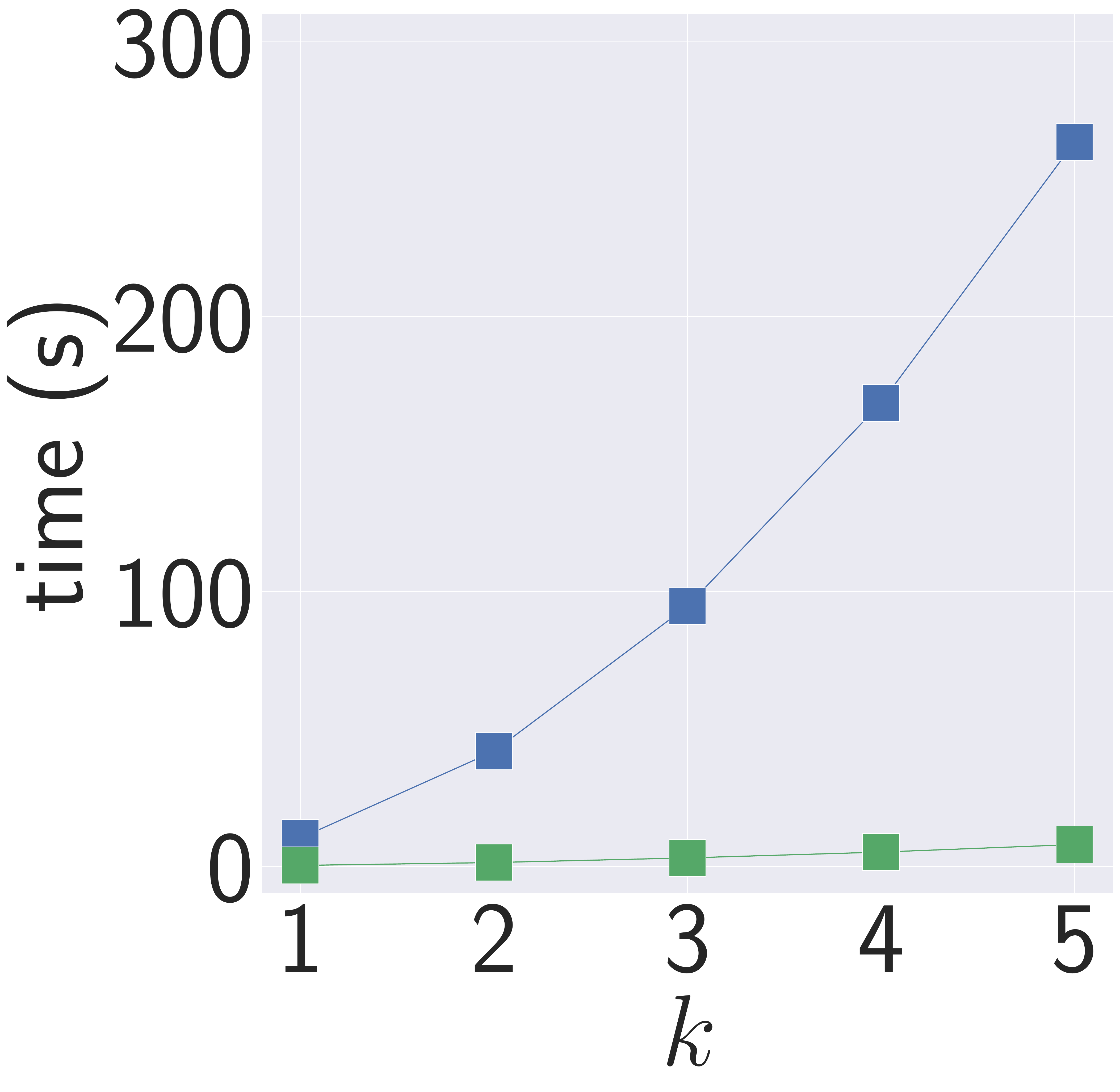} \label{fig:time_2}}
	\caption{Comparative time performance (measured on \DSMARQUES and \DSISRI). The difference between Marques and the proposed method is better noticed in (b). Despite Marques is very efficient, it does not deliver accurate solutions. The proposed approach is a feasible alternative since it reaches accurate solutions and is significantly more scalable than Paix\~ao and Liang.}
	\label{fig:time}
\end{figure}

Besides reaching better accuracy, the proposed approach is also remarkably more scalable in terms of time performance than Paix\~ao and Liang, as seen in \fig{fig:time}. Time scalability is a critical issue in real scenarios because it is expected much more than 5 shredded pages to be reconstructed. Note, in particular, that the time performance of Liang was the worst even leveraging heavy parallelism (240 threads). This is due to the overhead introduced by successive calls to the OCR software, which is the core of their method. Conversely, Marques is very time efficient, but, as shown in \fig{fig:comparison}, it delivered low accuracy reconstruction. Nonetheless, Marques' time performance will serve as the lower bound of efficiency for future optimizations of the proposed method.

\section{Conclusion}
\label{sec:conclusion}

This paper addressed the reconstruction of mixed text documents focusing on the central problem of evaluating the compatibility between shreds. The proposed segmentation-free deep learning approach enabled faster and more robust reconstruction of strip-shredded documents in more realistic scenarios and it also has the benefit of self-supervised learning, which facilitates scaling the training data.

To enable a better and more extensive evaluation, we introduced a new dataset containing 100 mechanically-shredded documents ($2{,}292$ shreds) with diverse layout. Despite the challenging scenarios, real-world cross-database experiments showed that our method achieved average accuracy superior to $90\%$ for different quantities of mixed documents. Nevertheless, the absence or scarcity of some patterns may hamper the proper reconstruction of the documents. A possible way to solve this problem is fine-tuning the model with samples from the inner region of the shreds belonging to the test documents themselves.

The ablation study evidenced that small and local samples are more effective for learning the compatibility between shreds. It is important, though, to consider this result in view of the limited diversity of the training data produced from relatively -- in the context of deep learning -- few documents. Additionally, the study showed the relevance of treating the misalignment between shreds at test time. An alternative approach for this issue is augmenting training data by simulating vertical misalignment. This would save processing time during the on-line reconstruction stage but could increase the complexity of the problem.

Comparative experiments showed that the accuracy of the proposed method (even in the modified version) was superior to the current state-of-the-art. When compared to Paix\~ao \cite{paixaotifs2019}, for instance, our method generalized better for documents with a more diverse layout and appearance, and also scaled more time-efficiently for the multi-document scenario. Furthermore, the time savings obtained by Marques \cite{marques2013} (based on the naive dissimilarity between edge pixels) were shown at the price of low reconstruction accuracy. Finally, the recently published Liang method \cite{liang2019reassembling} performed significantly worse than the proposed method in terms of accuracy, in addition to a limited time-scalability to real-world scenarios comprising several documents.

In addition to the mentioned directions, there are still other open problems to be addressed in our future research. First, reconstruction should be extended to even more realistic scenarios in which some shreds are missing, rotated, and/or significantly damaged (e.g., wet, torn, or wrinkled). Finally, we will also investigate the use of deep learning to extract text information (e.g, OCR, text/word spotting) from the reconstructed documents.
\section*{Conflict of interest}
The authors declare that they have no conflict of interest.
\section*{Acknowledgements}

This  study  was  financed  in  part  by  the Coordena\c{c}\~{a}o de Aperfei\c{c}oamento de Pessoal de N\'{i}vel Superior - Brasil (CAPES) - Finance Code 001.
Cloud computing resources were provided by AWS Cloud Credits for Research program. We thank the NVIDIA Corporation for 
the donation of a Titan Xp GPU used in this research. We also acknowledge the scholarships of Productivity on Research (grants 311120/2016-4 and 311504/2017-5) supported by Conselho Nacional de Desenvolvimento Cient\'{i}fico e Tecnol\'{o}gico (CNPq, Brazil).


\biboptions{sort&compress}

\begin{thebibliography}{10}
\expandafter\ifx\csname url\endcsname\relax
  \def\url#1{\texttt{#1}}\fi
\expandafter\ifx\csname urlprefix\endcsname\relax\def\urlprefix{URL }\fi
\expandafter\ifx\csname href\endcsname\relax
  \def\href#1#2{#2} \def\path#1{#1}\fi

\bibitem{cheng2019deep}
E.-J. Cheng, K.-P. Chou, S.~Rajora, B.-H. Jin, M.~Tanveer, C.-T. Lin, K.-Y.
  Young, W.-C. Lin, M.~Prasad, Deep sparse representation classifier for facial
  recognition and detection system, Pattern Recognit. Letters 125 (2019)
  71--77.

\bibitem{gomez20183d}
O.~G{\'o}mez, O.~Ib{\'a}{\~n}ez, A.~Valsecchi, O.~Cord{\'o}n, T.~Kahana, {3D-2D
  silhouette-based image registration for comparative radiography-based
  forensic identification}, Pattern Recognit. 83 (2018) 469--480.

\bibitem{li2020cvpr}
Y.~Li, X.~Yang, P.~Sun, H.~Qi, S.~Lyu, {Celeb-DF: A Large-Scale Challenging
  Dataset for DeepFake Forensics}, in: Conf. Comput. Vision and Pattern
  Recognit., 2020.

\bibitem{qureshi2019hyperspectral}
R.~Qureshi, M.~Uzair, K.~Khurshid, H.~Yan, Hyperspectral document image
  processing: Applications, challenges and future prospects, Pattern Recognit.
  90 (2019) 12--22.

\bibitem{he2019deep}
S.~He, L.~Schomaker, Deep adaptive learning for writer identification based on
  single handwritten word images, Pattern Recognit. 88 (2019) 64--74.

\bibitem{soleimani2016deep}
A.~Soleimani, B.~N. Araabi, K.~Fouladi, Deep multitask metric learning for
  offline signature verification, Pattern Recognit. Letters 80 (2016) 84--90.

\bibitem{townshend2019end}
R.~Townshend, R.~Bedi, P.~Suriana, R.~Dror, {End-to-End Learning on 3D Protein
  Structure for Interface Prediction}, in: Advances in Neural Information
  Processing Systems, 2019, pp. 15616--15625.

\bibitem{ostertag2020matching}
C.~Ostertag, M.~Beurton-Aimar, Matching ostraca fragments using a siamese
  neural network, Pattern Recognit. Letters 131 (2020) 336--340.

\bibitem{paixaotifs2019}
T.~M. Paix\~ao, M.~C.~S. Boeres, C.~O.~A. Freitas, T.~Oliveira-Santos,
  Exploring {Character} {Shapes} for {Unsupervised} {Reconstruction} of
  {Strip}-shredded {Text} {Documents}, IEEE Trans. Inf. Forensics Secur. 14~(7)
  (2019) 1744--1754.

\bibitem{perl2011}
J.~Perl, M.~Diem, F.~Kleber, R.~Sablatnig, Strip shredded document
  reconstruction using optical character recognition, in: Int. Conf. on Imag.
  for Crime Detection and Prevention, 2011, pp. 1--6.

\bibitem{badawy2018discrete}
H.~Badawy, E.~Emary, M.~Yassien, M.~Fathi, Discrete grey wolf optimization for
  shredded document reconstruction, in: Int. Conf. on Adv. Intell. System and
  Information, 2018, pp. 284--293.

\bibitem{marques2013}
M.~Marques, C.~Freitas, Document decipherment-restoration: Strip-shredded
  document reconstruction based on color, IEEE Latin America Trans. 11~(6)
  (2013) 1359--1365.

\bibitem{gong2016}
Y.-J. Gong, Y.-F. Ge, J.-J. Li, J.~Zhang, W.~Ip, A splicing-driven memetic
  algorithm for reconstructing cross-cut shredded text documents, Applied Soft
  Computing 45 (2016) 163--172.

\bibitem{chen2018high}
J.~Chen, D.~Ke, Z.~Wang, Y.~Liu, A high splicing accuracy solution to
  reconstruction of cross-cut shredded text document problem, Multimedia Tools
  and Appl. 77~(15) (2018) 19281--19300.

\bibitem{paixao2018deep}
T.~M. Paix\~ao, R.~F. Berriel, M.~C. Boeres, C.~Badue, A.~F. De~Souza,
  T.~Oliveira-Santos, A deep learning-based compatibility score for
  reconstruction of strip-shredded text documents, in: Conf, on Graphics,
  Patterns and Images, 2018, pp. 87--94.

\bibitem{morandell2008}
W.~Morandell, Evaluation and reconstruction of strip-shredded text documents,
  Master's thesis, Institute of Computer Graphics and Algorithms, Vienna
  University of Technology (2008).

\bibitem{prandtstetter2008}
M.~Prandtstetter, G.~R. Raidl, Combining forces to reconstruct strip shredded
  text documents, in: Hybrid Metaheuristics, Springer, 2008, pp. 175--189.

\bibitem{sleit2013}
A.~Sleit, Y.~Massad, M.~Musaddaq, An alternative clustering approach for
  reconstructing cross cut shredded text documents, Telecommunication Systems
  52~(3) (2013) 1491--1501.

\bibitem{phienthrakul2015}
T.~Phienthrakul, T.~Santitewagun, N.~Hnoohom, {A Linear Scoring Algorithm for
  Shredded Paper Reconstruction}, in: Int. Conf. on Signal-Image Tech. \&
  Internet-Based Syst., 2015, pp. 623--627.

\bibitem{smet2005}
P.~De~Smet, J.~De~Bock, W.~Philips, Semi-automatic reconstruction of
  strip-shredded documents, in: SPIE Electronic Imaging, Vol. 5685, 2005, pp.
  239--249.

\bibitem{skeoch2006}
A.~Skeoch, An investigation into automated shredded document reconstruction
  using heuristic search algorithms, Unpublished Ph. D. Thesis in the
  University of Bath, UK (2006) 107.

\bibitem{chen2017a}
G.~Chen, J.~Wu, C.~Jia, Y.~Zhang, {A pipeline for reconstructing cross-shredded
  English document}, in: IEEE Int. Conf. on Image, Vision and Computing, 2017,
  pp. 1034--1039.

\bibitem{chen2019solution}
J.~Chen, M.~Tian, X.~Qi, W.~Wang, Y.~Liu, {A solution to reconstruct cross-cut
  shredded text documents based on constrained seed K-means algorithm and ant
  colony algorithm}, Expert Syst. with Appl. 127 (2019) 35--46.

\bibitem{pomeranz2011}
D.~Pomeranz, M.~Shemesh, O.~Ben-Shahar, A fully automated greedy square jigsaw
  puzzle solver, in: IEEE Conf. Comput. Vision and Pattern Recognit., 2011, pp.
  9--16.

\bibitem{andalo2017}
F.~A. Andal\'o, G.~Taubin, S.~Goldenstein, {PSQP}: Puzzle solving by quadratic
  programming, IEEE Trans. on Pattern Anal. and Mach. Intell. 39~(2) (2017)
  385--396.

\bibitem{nartker2005}
T.~A. Nartker, S.~V. Rice, S.~E. Lumos, {Software tools and test data for
  research and testing of page-reading OCR systems}, in: Electronic Imag.,
  2005, pp. 37--47.

\bibitem{balme2007}
J.~Balme, Reconstruction of shredded documents in the absence of shape
  information, Tech. rep., Working paper, Dept. of Computer Science, Yale
  University, USA (2007).

\bibitem{ranca2013}
R.~Ranca, A modular framework for the automatic reconstruction of shredded
  documents, in: Workshops 27th AAAI Conf. on Artif. Intell., 2013.

\bibitem{lin2012}
H.-Y. Lin, W.-C. Fan-Chiang, Reconstruction of shredded document based on image
  feature matching, Expert Syst. with Appl. 39~(3) (2012) 3324--3332.

\bibitem{pohler2015}
D.~P{\"o}hler, R.~Zimmermann, B.~Widdecke, H.~Zoberbier, J.~Schneider,
  B.~Nickolay, J.~Kr{\"u}ger, Content representation and pairwise feature
  matching method for virtual reconstruction of shredded documents, in: Int.
  Symp. on Image and Signal Process. and Anal., 2015, pp. 143--148.

\bibitem{xing2017a}
N.~Xing, S.~Shi, Y.~Xing, {Shreds Assembly Based on Character Stroke Feature},
  Procedia Comput. Sci. 116 (2017) 151--157.

\bibitem{xing2017b}
N.~Xing, J.~Zhang, {Graphical-character-based shredded Chinese document
  reconstruction}, Multimedia Tools and Appl. 76~(10) (2017) 12871--12891.

\bibitem{liang2019reassembling}
Y.~Liang, X.~Li, {Reassembling Shredded Document Stripes Using Word-path Metric
  and Greedy Composition Optimal Matching Solver}, IEEE Trans. on Multimedia
  22~(5) (2020) 1168--1181.

\bibitem{le2018jigsawnet}
C.~Le, X.~Li, {JigsawNet: Shredded Image Reassembly using Convolutional Neural
  Network and Loop-based Composition}, arXiv preprint arXiv:1809.04137.

\bibitem{paumard2018jigsaw}
M.-M. Paumard, D.~Picard, H.~Tabia, {Jigsaw Puzzle Solving Using Local Feature
  Co-Occurrences in Deep Neural Networks}, in: IEEE Int. Conf. on Image
  Process., 2018, pp. 1018--1022.

\bibitem{doersch2015unsupervised}
C.~Doersch, A.~Gupta, A.~A. Efros, Unsupervised visual representation learning
  by context prediction, in: IEEE Int. Conf. on Comput. Vision, 2015, pp.
  1422--1430.

\bibitem{noroozi2016unsupervised}
M.~Noroozi, P.~Favaro, Unsupervised learning of visual representations by
  solving jigsaw puzzles, in: European Conf. on Comput. Vision, 2016, pp.
  69--84.

\bibitem{sholomon2016dnn}
D.~Sholomon, O.~E. David, N.~S. Netanyahu, {DNN-Buddies: a deep neural
  network-based estimation metric for the jigsaw puzzle problem}, in: Int.
  Conf. on Art. Neural Networks, 2016, pp. 170--178.

\bibitem{sauvola2000adaptive}
J.~Sauvola, M.~Pietik{\"a}inen, Adaptive document image binarization, Pattern
  Recognit. 33~(2) (2000) 225--236.

\bibitem{iandola2016squeezenet}
F.~N. Iandola, S.~Han, M.~W. Moskewicz, K.~Ashraf, W.~J. Dally, K.~Keutzer,
  {SqueezeNet: AlexNet-level accuracy with 50x fewer parameters and $<$0.5MB
  model size}, arXiv preprint arXiv:1602.07360.

\bibitem{kingma2014adam}
D.~P. Kingma, J.~Ba, {Adam: A Method for Stochastic Optimization}, in: Int.
  Conf. for Learn. Representations, 2015.

\bibitem{jonker1983}
R.~Jonker, T.~Volgenant, Transforming asymmetric into symmetric traveling
  salesman problems, Operations Research Letters 2~(4) (1983) 161--163.

\bibitem{applegate2003}
D.~Applegate, R.~Bixby, V.~Chvatal, W.~Cook, Concorde: {A} code for solving
  traveling salesman problems, \url{http://www.math.uwaterloo.ca/tsp/concorde},
  accessed on: June 10, 2020 (2003).

\bibitem{harley2015icdar}
A.~W. Harley, A.~Ufkes, K.~G. Derpanis, Evaluation of deep convolutional nets
  for document image classification and retrieval, in: Proc. IEEE Int. Conf. on
  Document Analysis and Recognit., 2015, pp. 991--995.

\bibitem{lewis2006building}
D.~Lewis, G.~Agam, S.~Argamon, O.~Frieder, D.~Grossman, J.~Heard, Building a
  test collection for complex document information processing, in: Conf. on
  Research and Develop. in Inf. Retrieval, 2006, pp. 665--666.

\end{thebibliography}

 \newcommand{\noop}[1]{}

\end{document}